\documentclass[pdflatex,sn-mathphys-num-nc]{sn-jnl}

\usepackage{graphicx}   
\usepackage{amsmath,amssymb,amsthm}
\usepackage{booktabs,multirow}
\usepackage{url}

\usepackage{comment}
\usepackage{tabularx}

\theoremstyle{plain}

\newtheorem{theorem}{Theorem}
\newtheorem{proposition}[theorem]{Proposition}

\newtheorem{example}{Example}
\newtheorem{remark}{Remark}

\newtheorem{definition}{Definition}

\title[Article Title]{Time-multiplexed layer reuse for physical neural networks}

\begin{document}

\author[1]{\fnm{Kohei} \sur{Tsuchiyama}}\email{tsuchiyama-kohei655@g.ecc.u-tokyo.ac.jp}

\author[1]{\fnm{Takatomo} \sur{Mihana}}\email{takatomo\_mihana@ipc.i.u-tokyo.ac.jp}

\author[1]{\fnm{Ryoichi} \sur{Horisaki}}\email{horisaki@g.ecc.u-tokyo.ac.jp}

\author*[1]{\fnm{Andr\'{e}} \sur{R\"{o}hm}}\email{roehm@g.ecc.u-tokyo.ac.jp}

\affil[1]{\orgdiv{Graduate School of Information Science and Technology}, \orgname{The University of Tokyo}, \orgaddress{\street{7-3-1 Hongo}, \city{Bunkyo-ku}, \postcode{113-8656}, \state{Tokyo}, \country{Japan}}}

\abstract{Physical neural networks (PNNs) are promising candidates for next-generation computing, but existing demonstrations remain several orders of magnitude smaller than modern digital neural networks, whose recent advances have been driven by rapid growth in trainable parameters. 
This situation resembles the constraints of early digital neural networks, which led to ideas around parameter reuse. 
We investigate what similarly efficient hardware architectures may look like, focusing specifically on the common bottleneck of slow re-adjustment of the weights in PNNs.
We propose the Time-Indexed Deep Alternating Layers Network (TIDAL-Net), which occupies an intermediate regime between recurrent and deep neural networks, specifically aimed at the scales and restrictions of common PNN prototypes. 
TIDAL-Net leverages the timescale separation found in many PNNs between fast forward dynamics and slowly trainable weights and biases, using layer-by-layer time multiplexing to increase effective depth while limiting implementation cost. 
Numerical experiments on image classification and natural language processing tasks show that TIDAL-Net improves performance with only minor modifications to conventional PNNs.}


\maketitle

\section*{Introduction}\label{sec1}
Artificial Intelligence (AI) technologies have made remarkable strides in areas such as image recognition \cite{lecun2015deep,simonyan2014very,redmon2016you} and natural language processing \cite{Transformer,devlin2019bert}.
Demand for computational power for AI is surging.
Until now, this demand has to be met by newer and more powerful data centers, based on semiconductor technologies.
However, these technologies are expected to increasingly face fundamental physical limitations in the near future.
One of the major concerns is the surge in power consumption as clock frequencies reach beyond gigahertz levels, resulting in excessive heat generation. 
Meanwhile, miniaturization is reaching scales where quantum effects begin to dominate. 
In response, researchers are actively exploring alternative avenues, from novel materials and architectures to the investigation of new computational paradigms.

A Physical Neural Network (PNN) describes the concept of implementing a neural network with technologies on other physical substrates such as optics \cite{wetzstein2020inference, appeltant2011information, vandoorne2014experimental, shen2017deep, tait2017neuromorphic, bueno2018reinforcement, lin2018all, ramey2020silicon,
onodera2025arbitrary,
jung2022crossbar,
wright2022deep,
xia2023hardware,
nakajima2022physical,
kalinin2025analog}, spintronics \cite{torrejon2017neuromorphic, Plouet2025, romera2018vowel}, and chemitronics \cite{parrilla2020programmable, kan2022physical}.
PNNs are often, but not necessarily, analog, and they may employ setups that do not separate memory and computation, thus differing from traditional von Neumann architectures.
PNNs ultimately aim to supplement and surpass conventional silicon-transistor systems within the specific field of artificial neural networks, either in parts or end-to-end, by exploiting new paradigms and physical sources of computation. 

\begin{figure}[h]
\centering
\includegraphics[width=\textwidth]{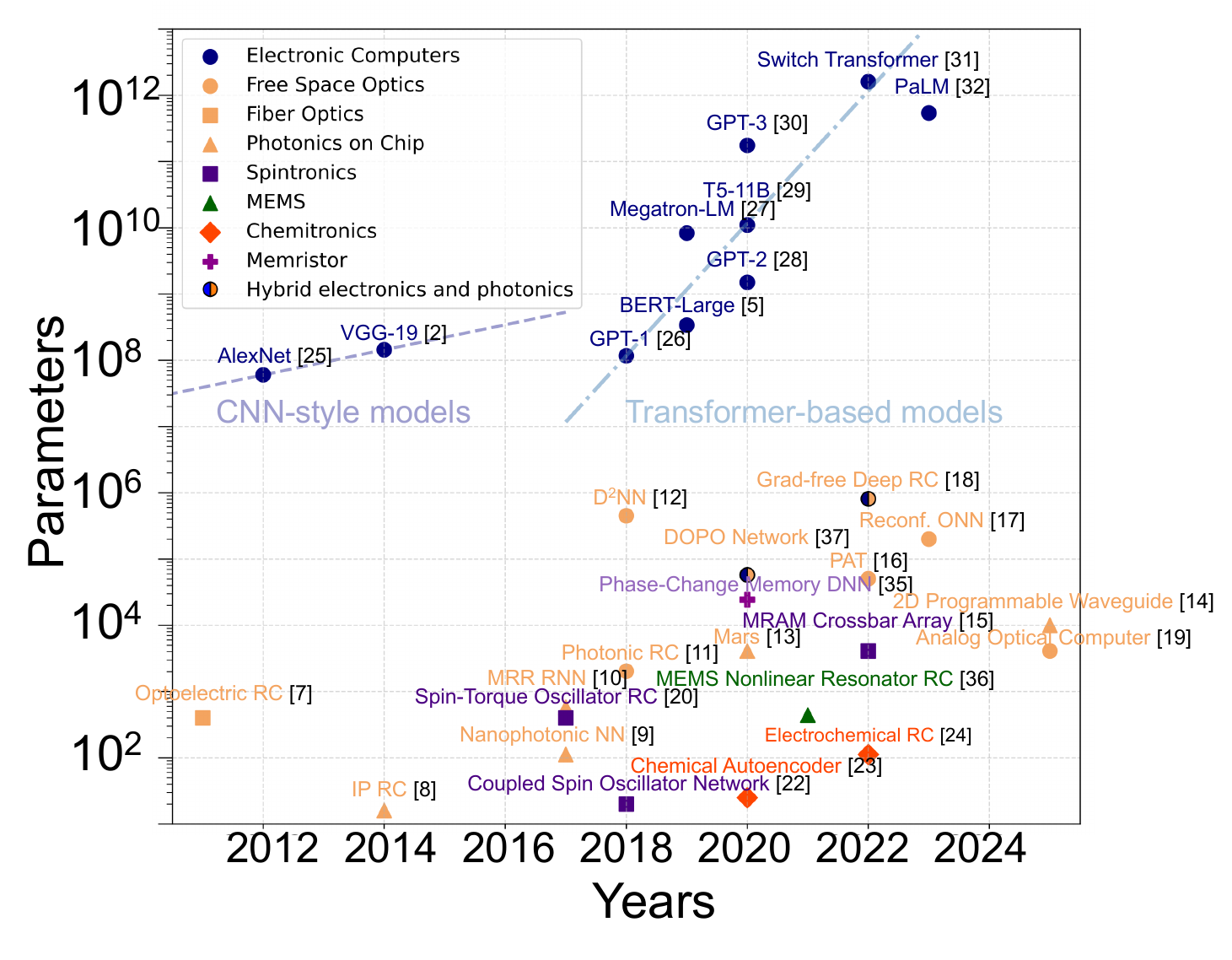}
\caption{Representative reported model scales of neural networks implemented on conventional digital electronic computers and of physical neural networks (PNNs). Reference numbers are shown next to the corresponding data points, and marker colors and shapes indicate the implementation platforms. For digital models, the plotted value is the number of trainable scalar parameters reported by the original authors or developers. For PNNs, however, a unique parameter count that is independent of the architecture and physical platform is generally not available. We therefore use the reported or estimated number of trainable or programmable degrees of freedom when such a quantity can be identified. When no direct parameter count is defined, we use the closest platform-specific model-size proxy that can be derived from the original publication, such as the number of controllable physical elements, physical nodes, or virtual nodes. The exact quantity, source, and calculation used for each data point are provided in Supplemental Information Section S8. Consequently, the plotted quantity should not be interpreted as an architecture- or platform-independent measure of expressivity, representational capacity, task performance, or hardware efficiency. The figure is intended only as a qualitative comparison of the reported implementation scales of digital and physical neural-network systems.
}
\label{fig:FIGINTRO}
\end{figure}

Recent advances in AI performance are largely driven by the rapid growth in the number of trainable parameters of neural network models~\cite{krizhevsky2012imagenet,simonyan2014very,radford2018improving,devlin2019bert,shoeybi2019megatron,radford2019language,raffel2020exploring,brown2020language,fedus2022switch,chowdhery2023palm}.
Larger parameter counts enable more complex representations, mappings and transformations within the network, ultimately yielding a higher performance. 
For example, in Large Language Models (LLM), well-known scaling laws capture this relationship \cite{kaplan2020scaling}.
Therefore, to produce ever better LLMs, significant effort has been devoted to scaling up the size \cite{villalobos2022machine}. 
Thus, PNNs should also support a comparable number of parameters to be competitive with current digital systems.
However, the number of parameters in published and demonstrated PNNs is low and, arguably, stagnating (see Fig.~\ref{fig:FIGINTRO}) \cite{bueno2018reinforcement, lin2018all, wright2022deep, xia2023hardware, kalinin2025analog, appeltant2011information, vandoorne2014experimental, shen2017deep, tait2017neuromorphic, ramey2020silicon, onodera2025arbitrary, yao2020fully, nakajima2022physical, torrejon2017neuromorphic, romera2018vowel, jung2022crossbar, sun2021novel, parrilla2020programmable, kan2022physical, inagaki2021collective}. 
While comparing only the number of trainable parameters is not a perfect measure of expressivity, qualitatively we can assess that the gap between digital and PNNs is large and, seemingly, growing.
Where recent LLMs have sizes of hundreds of billions of trainable parameters \cite{brown2020language} (Fig.~\ref{fig:FIGINTRO}, blue dots), the largest PNN demonstrations do not even reach a million (Fig.~\ref{fig:FIGINTRO}, other colors).

Physical constraints inherent to non-electronic devices often pose significant challenges to scaling up layer-wise parameters in PNNs. 
For example, diffraction divergence and physical footprint limit integration density in free space optics \cite{hu2024diffractive}, thermal crosstalk and footprint-loss tradeoff restrict scaling in integrated photonic circuits \cite{al2022scaling, biasi2022effect}, and coupling overhead and device variability hinder enlarging in spintronics \cite{grollier2020neuromorphic}.
Overcoming these challenges typically demands large and complex architectures, resulting in unfavorable cost problems \cite{mcmahon2023physics}.
As a result, no existing PNN has yet matched the computational capacity of high-end electronic systems.

As the raw number of parameters of PNNs is likely to remain below that of electronic systems, we must investigate alternative forms of using the existing sizes effectively.
These constraints mirror the size restrictions found in early digital neural networks, and one can take inspiration from relevant papers of the era.
For example, the development of Convolutional Neural Networks (CNNs) was, in part, driven by a need to be parameter-efficient \cite{lecun1989handwritten}. 
Here, the common insight was the following: Even when the number of trainable weights is fixed, one can still increase the size of the neural network --- if one allows the re-use of components. 
This results in networks, where certain edge weights are forcibly set to always be identical to others. 
While this does reduce the number of trainable parameters, it still preserves a high dimensional operation - such as in the case of convolutional Kernels in CNNs \cite{lecun1989handwritten}.
Such ``weight-sharing'' or ``tying-weights'' is useful for engineering neural
networks towards specific tasks. 
Indeed, convolutional Kernels may be a viable path for increasing the power of PNNs, and first demonstrations exist that demonstrate efficient re-use of weights. 
For example, Fan et al. experimentally realized convolution processing in a photonic synthetic frequency dimension, using time modulation in a ring resonator \cite{fan2023experimental}.

In this paper, we aim to provide a framework for weight-sharing and hardware reuse for PNNs, taking inspiration from the historical precedents that digital neural networks underwent, but also taking into account the unique properties of the alternative hardware platforms under exploration.
We start from the insight that digital computers scale because they decouple parameters from hardware: a small, fixed set of multiply-accumulate units is reused - time-multiplexed - across all of a network's weights within a single forward pass, so the physical compute substrate stays small while the network it evaluates can be much larger.
Time-multiplexing is an important technique that enables scaling in time. 
Here, we must accommodate the unique properties of many PNN platforms: While digital computers have access to high-speed DRAM for storing a large number of weights, physical substrates are often too slow for such weight reconfiguration. 

We present an intermediate model between Recurrent Neural Network (RNN) and Deep Neural Network (DNN) called Time-Indexed Deep Alternating Layers Network (TIDAL-Net) whose architecture increases the capability of PNNs. 
TIDAL-Net is based on weight-sharing via time-multiplexing.
It uses a small set of layers and switches between them at each calculation time step.
TIDAL-Net is motivated by combining the benefits of both RNNs and DNNs. 
When sufficiently large persistent parameter banks are available, TIDAL-Net can extend such systems by switching among these banks while reusing the nonlinear and state-update hardware. 
Future PNNs should try to maximize the reuse of as many elements as possible, so that there is no need to replicate nonlinear activation components or reconfigure the entire network at high rates.


Stepping back, we see the proposed TIDAL-Net framework as a model representing an intermediate-scale PNN, which is positioned between conventional small PNNs with static weights and the fully dynamic large PNNs of the future.
Just as ``Noisy Intermediate-Scale Quantum computing" (NISQ) \cite{preskill2018quantum} is seen as an intermediate step towards general use in quantum computing, the TIDAL-Net architecture is mainly motivated by the limitations of near-future PNNs in mind. 
To truly compete with digital silicon-transistor based digital computers, PNNs will need to become reconfigurable and time-multiplexed to a much higher degree than they are now.

\section*{Results}
\subsection*{Time-multiplexing of neural networks}
\label{sec:time_mulitplexing}

\begin{figure}[h]
\centering
\includegraphics[width=\textwidth]{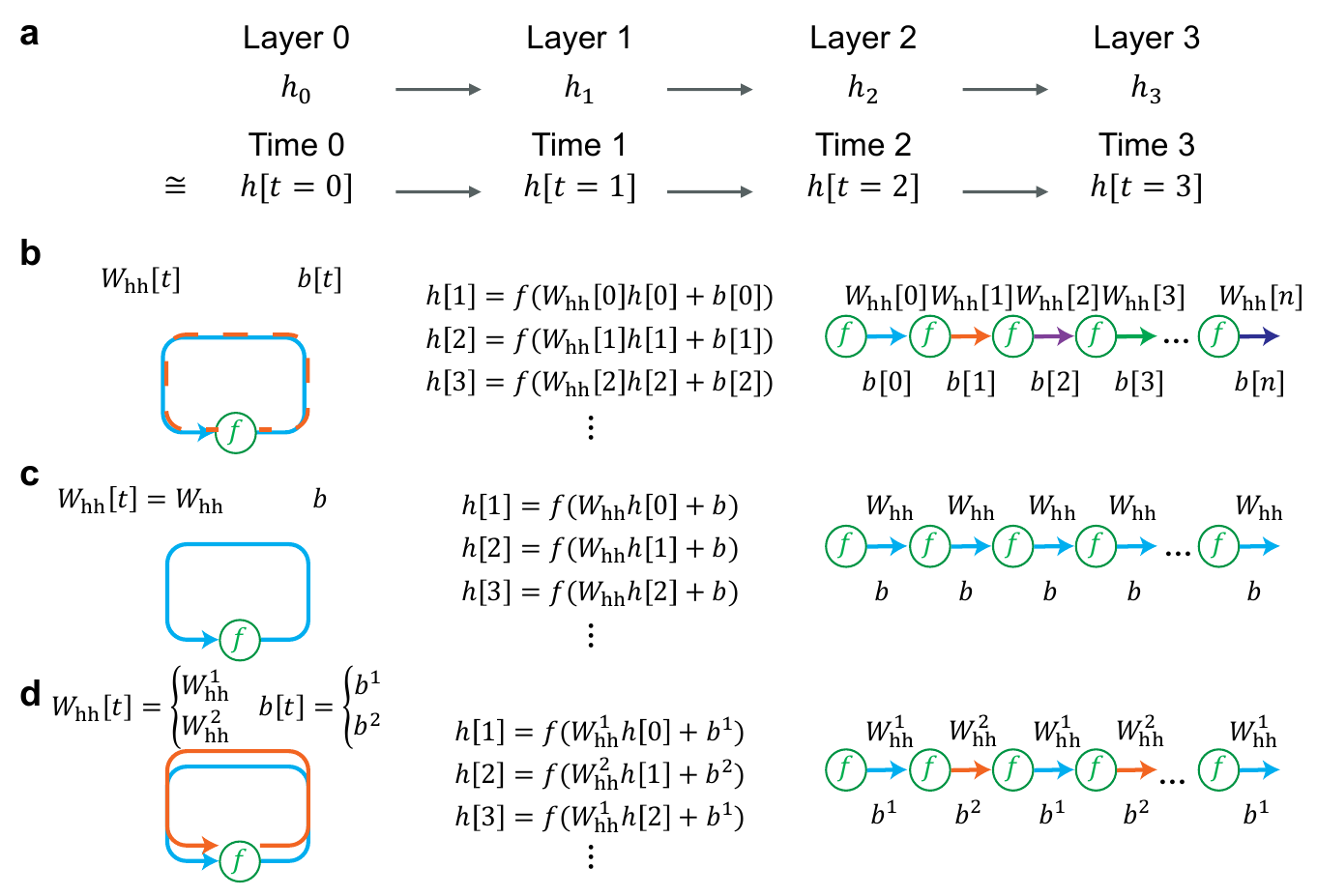}
\caption{Concepts of ``time-multiplexing". $\mathbf{a}$: While the state $h_l$ in a conventional network evolves when passing through distinct layers $l$, in a time-multiplexed system the state $h[t]$ evolves over several time steps $t$ (see Eq.~\eqref{eq:time-mutliplexed-DNN}). $\mathbf{b}$: By reusing the same nonlinear element, but reconfiguring the network weights $W_{\rm{hh}}[t]$ (and biases $b[t]$) at each step $t$, we can reproduce the dynamics of a DNN (Eq.~\eqref{eq:time-mutliplexed-DNN}).  $\mathbf{c}$: If the matrix $W_{\rm{hh}}$ (and bias) is fixed (independent of $t$), the system implements a RNN (Eq.~\eqref{eq:RNN}). $\mathbf{d}$: The network weight $W_{\rm{hh}}$ and biases $b[t]$ are switched periodically at every time step for the TIDAL-Net architecture proposed in Methods. 
}
\label{fig:time_multiplexing}
\end{figure}

Consider a multi-layer perceptron with $N$ neurons forming the hidden states $h_l \in \mathbb{R}^N$ for the layers $l \in \{0, \dots, L\}$ where $l=0$ corresponds to the input state. The forward pass for calculating states $h_l$ and final output $y$ via the nonlinear activation function $f$ is typically formulated as 
\begin{align}
h_{l+1} &= f (W_{\rm{hh}}^l  \, h_l + b_l), \label{eq:general_DNN}\\
\hat{y}&= W_{\rm{hy}}h_{L} + b_{\rm{y}}\label{eq:general_output}
\end{align}
where the entries of the hidden-hidden weight matrices $W_{\rm{hh}} \in \mathbb{R}^{N\times N}$, the hidden-output weight matrices $W_{\rm{hy}}^l \in \mathbb{R}^{Y\times N}$ and bias terms $b_l \in \mathbb{R}^N$, $b_{\rm{y}} \in \mathbb{R}^Y$ are to be determined via training.
We consider a constant hidden state dimension $N$ for all $l$ here for simplicity, but the argument can in principle be generalized.

The core challenge when creating Physical Neural Networks (PNNs) is to consider how to emulate this functionality in hardware.
The most straightforward way is to create a separate physical neuron for each dimension of $h$ and for each layer $l$, each implementing the nonlinear activation function $f$. 
These physical neurons then must be connected with adjustable weights to emulate the effects of the weight matrix $W_{\rm{hh}}^l$ for every layer $l$.
Thus, $N \times L$ dynamically changing variables are needed for simulating the evolution of the states $h_l$, as well as $N^2 \, L + N \, L$ trainable parameters for implementing the weights and biases.
This naive implementation scales poorly - larger networks will require ever larger hardware, where each new neuron and each new edge requires the corresponding hardware parts.
Thus, having such a one-to-one correspondence of physical components and neurons/weights is hardware-inefficient.

The idea of ``time-multiplexing" can drastically reduce the hardware need --- by trading spatial complexity for temporal complexity.
Time-multiplexing refers to techniques that reuse the same component several times, each time implementing a different part of the neural network calculation.
For example, instead of implementing $L$ parallel hardware implementations of the nonlinear activation function $f$, one per layer, it is more efficient to reuse it sequentially. 
In principle, if all the layers have the same width, the same physical nonlinearity $f$ can be used for a large amount in sequence.

A general but clean formulation of a time-multiplexed neural network in discrete time can be written as follows.
The hidden state $h[t] \in \mathbb{R}^N$ now evolves in \textit{time} $t \in [0, \dots, L_\mathrm{T}] $ via
\begin{align}
h[t+1] &= f (W_{\rm{hh}}[t] \, h[t] + b[t]), \label{eq:time-mutliplexed-DNN}\\
\hat{y}&= W_{\rm{hy}}h[L_{\rm{T}}] + b_{\rm{y}}. \label{eq:time-multiplicated_output}
\end{align}
By simply assigning $W_{\rm{hh}}[t] = W_{\rm{hh}}^{l=t}$ and $b[t] = b_{l=t}$, we can reproduce exactly the same dynamics as the original MLP formulated in Eq.~\eqref{eq:general_DNN}. 
As sketched by Fig.~\ref{fig:time_multiplexing}$\mathbf{a}$, the layers are now explicitly calculated sequentially.
Instead of information of distinct layers of the neural network being physically located in separate elements, we reuse the hidden state $h[t]$, the weights $W_{\rm{hh}}[t]$ and bias terms $b[t]$ for each time step. 

Looking closely at Eq.~\eqref{eq:time-mutliplexed-DNN}, we can quickly see that a single physical platform that can implement the right hand side of Eq.~\eqref{eq:time-mutliplexed-DNN} would in principle suffice to implement the entire depth of the neural network, independent of $L$.
However, this requires that $W_{\rm{hh}}[t]$ can be reconfigured fast enough, i.e., all its entries must be updated for each new time step $t$.
This kind of ``time-variable RNN" (see Fig.~\ref{fig:time_multiplexing}{\bf b}) utilizes ideas similar to Neural ODEs \cite{chen2018neural} and has also been called ``Time-varying Recurrent Neural Networks" \cite{zhang2025behavioral} in the literature.
In principle, we can even time-multiplex all the neurons within each layer $l$, reducing $h[t]$ to a scalar, further drastically lowering the number of necessary distinct physical elements, as shown in ``Folded-in-time Deep Neural Networks" \cite{stelzer2021deep}. 
However, this requires additional trade-offs in terms of connectivity and implementation complexity, so in this manuscript we consider that $h[t]$ physically stores the state of an entire hidden layer.

Reconfiguring the network weights $W_{\rm{hh}}[t]$ and bias terms $b[t]$ at each step \cite{ha2016hypernetworks, hughes2019wave, ba2016using} is an effective approach for creating larger neural networks with fewer hardware components. 
Indeed, we should think of classical digital neural networks as heavily employing this strategy.
The parameters and activations of digital neural networks are usually intermittently stored and loaded in the cache memory close to the processors. 
Thus, the matrix-vector products and updates for forward passes are in actuality calculated sequentially even in digital computers. 
Note, that this is only worthwhile, because digital computers can effectively load new values of these parameters at high speeds.

Many PNNs do not have the ability to reconfigure their parameters fast enough to use such a complete time-multiplexing strategy.
The physical mechanisms used for training and setting the weight parameter values are typically slower compared to the dynamical timescales of forward inference, as discussed in the Supplemental Information Section S10. 
Creating PNNs with quickly reconfigurable weights is indeed a promising avenue for the field \cite{shen2017deep,tait2017neuromorphic,jung2022crossbar,grollier2020neuromorphic,tait2018feedback}.
But for those PNNs with parameters that cannot be changed on single-pass timescales, we must consider alternative ways of time-multiplexing.

In the extreme case, where we cannot reconfigure the weights during inference at all in Eq.~\eqref{eq:time-mutliplexed-DNN}, we can only hope to re-use the same layer over and over:
\begin{align}
h[t+1] &= f (W_{\rm{hh}} \, h[t] + b), \label{eq:RNN}
\end{align}
which describes the case of $W_{\rm{hh}} = \rm{const}$ resulting in a conventional ``stateless RNN" \cite{graves2013generating} (Fig.~\ref{fig:time_multiplexing}{ \bf c}). 
This choice does not limit the number of ``layers" we can implement --- we merely need to repeat the same operation over and over, if we so desire, although typical RNN language would not refer to the number of repetitions as layers.
In the limit of infinitely many repetitions, this approaches a deep equilibrium model \cite{bai2019deep}.
The number of trainable parameters is however limited to the dimensions of $W_{\rm{hh}}$ and $b$.

At this stage, one may think that these well-known options are the only ones available to a PNN --- either parameters can be reconfigured fast, or they cannot. 
If, in line with realities of most PNNs, the parameters of the weight matrices are assumed to be fixed during forward-passes, does this imply our only choice is to implement a classical stateless RNN, or to pay the high hardware cost of having a separate hardware element for each layer? 
In fact, there is an additional class of time-multiplexed networks that lies conceptually in-between the stateless RNN of Eq.~\eqref{eq:RNN} (Fig.~\ref{fig:time_multiplexing}~\textbf{c}) and the fully time-variable DNN of Eq.~\eqref{eq:time-mutliplexed-DNN} (Fig.~\ref{fig:time_multiplexing}~\textbf{b}), both of which are already widely used.
This is the idea behind the proposed TIDAL-Net architecture (Fig.~\ref{fig:time_multiplexing}{\textbf{d}}), which is explained in detail in the following section.

\subsection*{Time-Indexed Deep Alternating Layers Network (TIDAL-Net)}\label{sec:TIDAL-Net}

\begin{figure}[h]
\centering
\includegraphics[width=\textwidth]{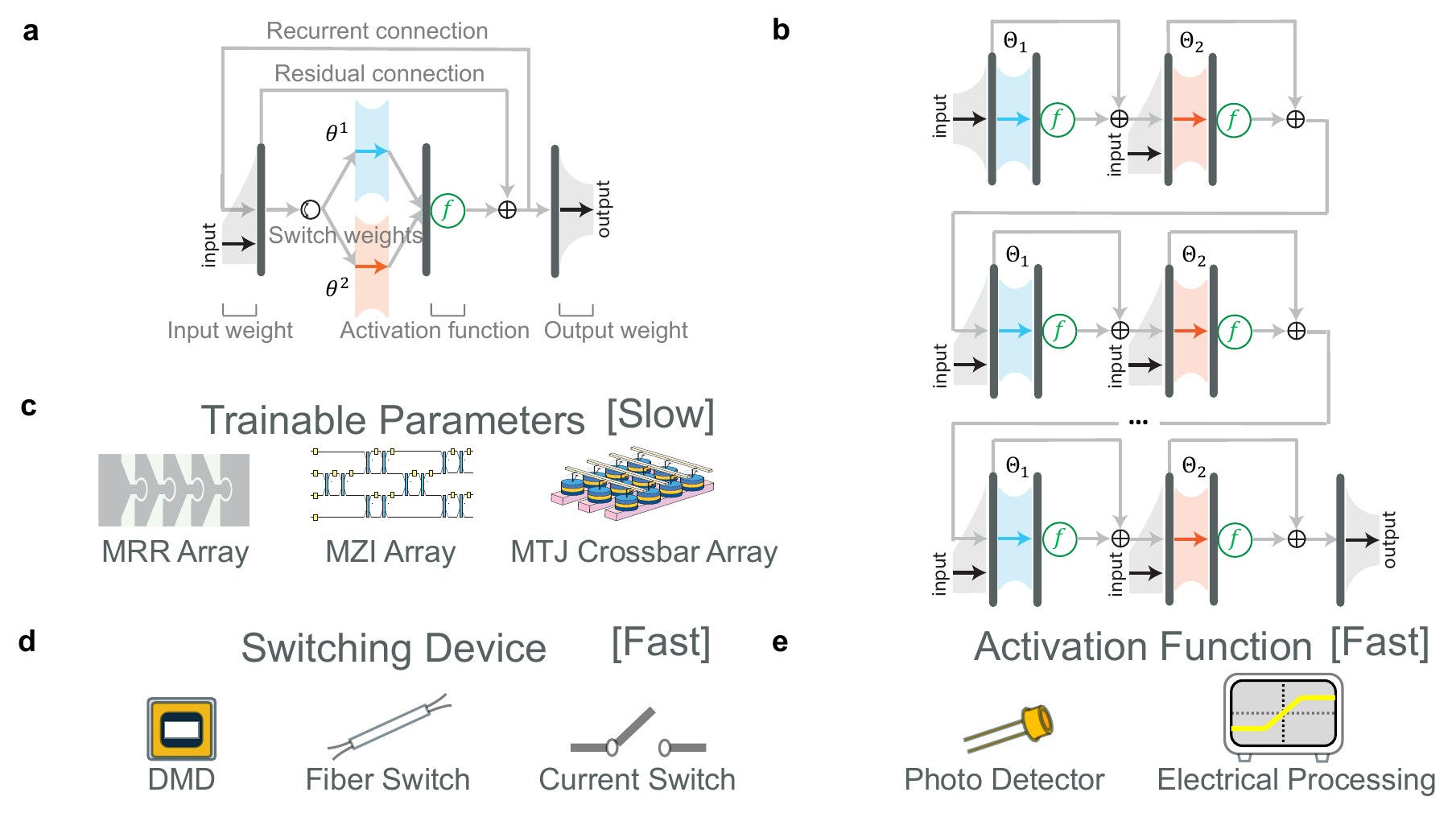}
\caption{Time-Indexed Deep Alternating Layers Network (TIDAL-Net). \textbf{a}: The forward propagation switches between separate sets of hidden layers with different parameters $\theta=(W_{\rm{hh}},b_{\rm{h}})$, see Eq.~\eqref{eq:relax_net}. 
This enables the hardware-efficient reuse of many components while only requiring high-speed switches (parameter reconfiguration can happen at much slower speeds).
\textbf{b}: If we consider how layers are chained temporally, this forms a deep neural network.  
The total depth is given by the number of repetitions $L_{\rm{T}}$.
\textbf{c, d, e}: Examples of proposed implementations for PNN components:
\textbf{c}: Micro-Ring Resonator (MRR), Mach-Zehnder Interferometer (MZI), and Magnetic Tunnel Junction (MTJ) crossbar arrays for trainable parameters;
\textbf{d}: Digital Mirror Devices (DMDs) \cite{hornbeck1997digital}, fiber switches \cite{friberg1988femotosecond}, and current switches for switching;
\textbf{e}: photodetectors or electrical processing for the nonlinear activation function.
}
\label{fig:relax_net_PNN}
\end{figure}

We consider new ways of implementing multilayer PNNs to relax physical constraints, while improving the performance.
Specifically, we expand upon the idea of simple stateless RNNs as illustrated in Fig.~\ref{fig:time_multiplexing}{$\textbf{c}$}. 
The main conceptual hurdle, as identified in the previous section, is that we typically cannot assume that the weights $W_{\rm{hh}}[t]$ and bias parameters $b[t]$ can be reconfigured quickly, making it infeasible to construct a neural network via layer-by-layer time-multiplexing (see Eq.~\eqref{eq:time-mutliplexed-DNN} and Fig.~\ref{fig:time_multiplexing}{\textbf{b}}).
Instead, under some reasonable assumptions, we can explore neural network architectures that are still plausibly constructible with such restricted PNN hardware, but exceed stateless RNNs in hardware efficiency. 

The main idea is the following:
While the parameters $\theta_l = \{W_{\rm{hh}}[t], b[t] \}$ of each physical layer themselves may be fixed, fast switches can still allow PNNs to be constructed flexibly from multiple fixed layers. 
For example, in the simplest case, we may assume two physically implemented weight matrices $W_{\rm{hh}}^0$ and $W_{\rm{hh}}^1$.
By chaining the first and second kind of weight matrix, we can create a non-trivial deep structural network, with a different functionality than a stateless RNN.
These deep structural networks are described by a certain structure of tying-weights --- certain layers sharing the same parameters --- which allows us to explore a richer set of architectures.
Tying weights has been studied for digital models, and remains an active field within transformers \cite{takase2023lessons,bae2025relaxed}.

The TIDAL-Net architecture is a residual neural network with $L_{\rm{T}}\in\mathbb{N}$ layers, described by the time evolution of the hidden state $h[t] \in \mathbb{R}^N$:
\begin{align}
    h[t+1] &= \alpha h[t] + f (W_{\rm{xh}}x[t] + W_{\rm{hh}}[t]h[t] + b_{\rm{h}}[t]), \label{eq:relax_net}
\end{align}
where $W_{\rm{xh}}$ describes the input weights for input $x[t]$ and $\alpha$ is the strength of the residual connection or skipping connection, which is adopted for stable learning.
Each time step $t$ corresponds to one of the layers $t = \{ 0, 1, \dots, L_{\rm{T}} \}$. 
The parameters of the hidden layers $\theta[t] = \{W_{\rm{hh}}[t], b_{\rm{h}}[t]\}$ change periodically as
\begin{align}
    \theta[t] &= 
    \begin{cases}
         \theta^1\ (t \mod L_{\rm{W}} \equiv0 )\\
         \theta^2\ (t \mod L_{\rm{W}} \equiv1 )\\
         ... \\
         \theta^{L_{\rm{W}}}\ (t \mod L_{\rm{W}} \equiv L_{\rm{W}} - 1 ) 
      \end{cases} 
\end{align}
with $L_{\rm{W}}$ being the number of different parameter sets for the hidden layers, i.e., weights and biases, which correspond to chaining weight matrices \cite{takase2023lessons}.
Thus, we decouple the number of layers in time $L_{\rm{T}}$ and the number of distinct hidden weight parameter banks $L_{\rm{W}}$.
Figure \ref{fig:relax_net_PNN} illustrates TIDAL-Net in the case of two periodically switched layers $L_{\rm{W}}=2$.

In hardware, the $L_\mathrm{W}$ trainable parameter banks must be physically instantiated, for example with MRR, MZI, or MTJ arrays (Fig.~\ref{fig:relax_net_PNN}\textbf{c}), while the nonlinear/state-update path can be reused.
Inference then requires fast routing among these banks and no forward-pass-rate reprogramming of their weights; representative switching and reconfiguration timescales are discussed below and in Supplemental Information Section S10.

\begin{figure}[h]
\centering
\includegraphics[width=\textwidth]{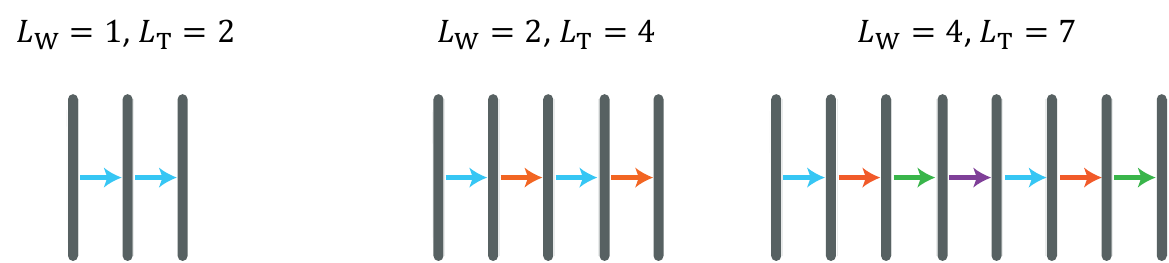}
\caption{Examples of TIDAL-Net. From left to right, the cases are: $(L_{\rm{W}}, L_{\rm{T}})= (1,2), (2,4), (4,7)$. The gray solid line represents the hidden states, while the colored arrows inbetween represent the parameters of the transformations (weights and biases). 
Shared colors indicate that parameters were reused, whereas differing colors indicate unique parameters. 
$L_{\rm{T}}$ corresponds to the number of layers and $L_{\rm{W}}$ corresponds to the number of distinct colors of the arrows, i.e., parameter sets.
}
\label{fig:L}
\end{figure}

By changing the time-multiplexing steps $L_{\rm{T}}$ and number of unique physical parameter sets  $L_{\rm{W}}$, we can implement a variety of different setups, as shown in Fig.~\ref{fig:L}, where different colors indicate distinct parameters $\theta^l$ for layers. 
In particular, the classical case of RNN is recovered by choosing $L_{\rm{T}} > 1$ and $L_{\rm{W}} = 1$ (see Fig.~\ref{fig:L}, left).
Conversely, choosing $L_{\rm{W}} = L_{\rm{T}}$ would mean that no layers are re-used, and we instead have recreated a direct correspondence between hardware and model parameters.
The hardware-efficient setups that we aim to explore involve all cases of $1 < L_{\rm{W}} < L_{\rm{T}} $, where at least some layers are reused, as shown on the right of Fig.~\ref{fig:L}.

\subsection*{TIDAL-Net performance on typical benchmark tasks}\label{section_performance_example}

We perform numerical experiments to demonstrate how the switching affects the number of trainable parameters, and how TIDAL-Net scales in terms of hardware efficiency. 
Firstly, we explore the scaling of the performance when going from a stateless RNN to a fully time-variable DNN, where TIDAL-Net describes the cases in between.
Then, we investigate the general behavior of TIDAL-Net for various values of $L_{\rm{T}}$ and $L_{\rm{W}}$ and how the model benefits from the temporal repetition. 
Finally, we investigate what choice of $L_{\rm{T}}$ is optimal under a fixed parameter budget.

We evaluated TIDAL-Net on two benchmark tasks: one each from image classification and Natural Language Processing (NLP). 
Details of the datasets, preprocessing, training procedure, and parameter counting are provided in the Methods section.
We first fixed the effective depth $L_\mathrm{T}$ and varied the number of distinct hidden-layer parameter banks $L_\mathrm{W}$, thereby interpolating between the stateless RNN limit and the untied multilayer limit.

\subsubsection*{Benchmark task performance}\label{subsection_task_performance}

\begin{figure}[h]
\centering
\includegraphics[width=\textwidth]{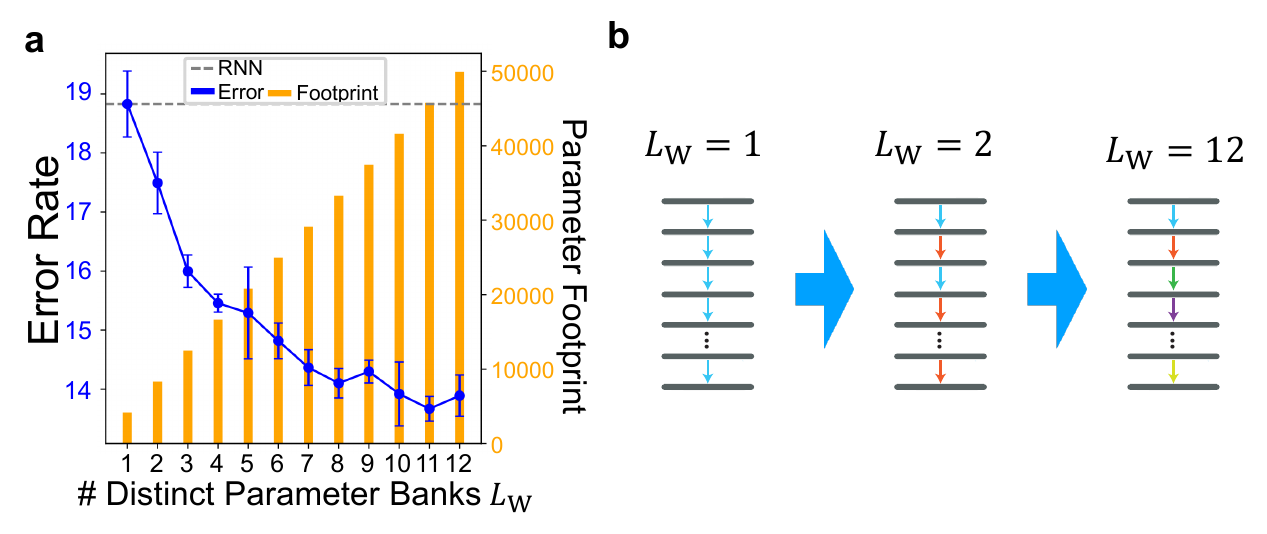}
\caption{TIDAL-Net performance on the image classification task. the number of distinct parameter banks $L_{\mathrm{W}}$ is changed while keeping the number of layers $L_{\mathrm{T}}=12$ fixed. 
$\mathbf{a}$: Error Rate and Parameter Footprint when varying $L_{\rm{W}}$. 
Blue dots and lines show the error rate, and orange bars indicate the hidden-layer parameter footprint defined in Methods.
The dashed line is the performance of the stateless RNN limit for $L_{\rm{W}} = 1$. 
$\mathbf{b}$: Illustration of TIDAL-Net architectures. Gray bars represent the hidden layers, and arrows indicate hidden weights. 
(Left) Reusing the same hidden weight. ($L_{\mathrm{W}}=1$)  (Middle) Repeating two weights. ($L_{\mathrm{W}}=2$) (Right) Using a distinct hidden parameter bank at every time step ($L_\mathrm{W}=12$)
TIDAL-Net $L_{\mathrm{W}}\geq2$ cases are better than RNN $L_{\mathrm{W}}=1$ cases. 
Especially, with a few additional layers, TIDAL-Net can outperform RNNs.
}
\label{fig:svhn_results}
\end{figure}

Figure~~\ref{fig:svhn_results} shows the image-classification results.
The error decreases steeply for the first few additional parameter banks and then
saturates.
Although increasing $L_\mathrm{W}$ also increases the number of trainable parameters,
the next subsection shows that temporal repetition itself contributes to the improvement.

\begin{figure}[h]
\centering
\includegraphics[width=0.5\textwidth]{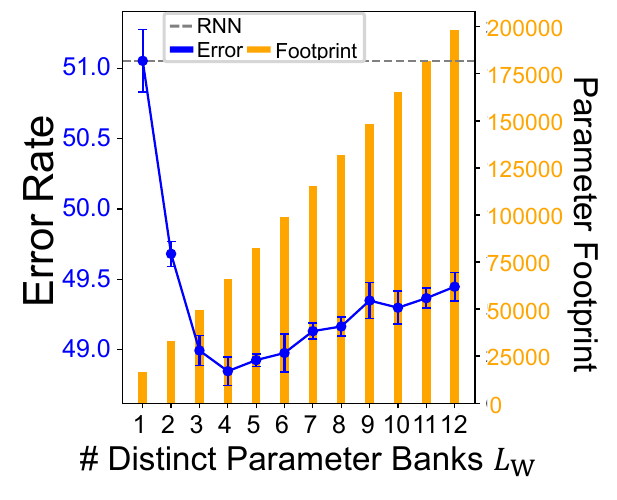}
\caption{TIDAL-Net performance on a natural language processing task. 
the number of distinct parameter banks $L_{\rm{W}}$ is changed while keeping the number of layers $L_{\rm{T}}=12$ constant. 
Blue dots and lines show the Error Rate, and orange bars show the number of trainable parameters, corresponding to the hardware footprint. 
The gray dashed line is the performance of the stateless RNN limit for $L_{\rm{W}} = 1$. 
Especially, with a few additional layers, TIDAL-Net can outperform RNNs.
}
\label{fig6:NLP}
\end{figure}

Figure~\ref{fig6:NLP} shows the result on the NLP task.
Here, the decline of the error (blue line) is even steeper with $L_{\rm{W}}$.
Performance does not increase when $L_{\rm{W}}$ reaches around $4$, which is likely caused by vanishing gradients.

For both tasks, we can confirm that the performance for $L_{\rm{W}}>1$ consistently surpasses the $L_{\rm{W}}=1$ RNN case.
These results confirm that TIDAL-Net achieves performance gains with only moderate modifications, supporting its feasibility for scalable physical implementations.
We have also confirmed that these results are consistent when using a different nonlinear activation function, especially on image classification, see the Supplemental Information Section S3 for details.

\subsubsection*{Importance of repetition in TIDAL-Net}\label{subsection_repetition}

One of TIDAL-Net's distinctive features is its controlled repetition of weights, bridging characteristics of both RNNs and DNNs.
Specifically, being different from a multilayer perceptron (DNN), TIDAL-Net reuses some weights like an RNN.
Here, we investigate how repetitive use of multiple weights affects the performance on TIDAL-Net.

\begin{figure}[h]
\centering
\includegraphics[width=\textwidth]{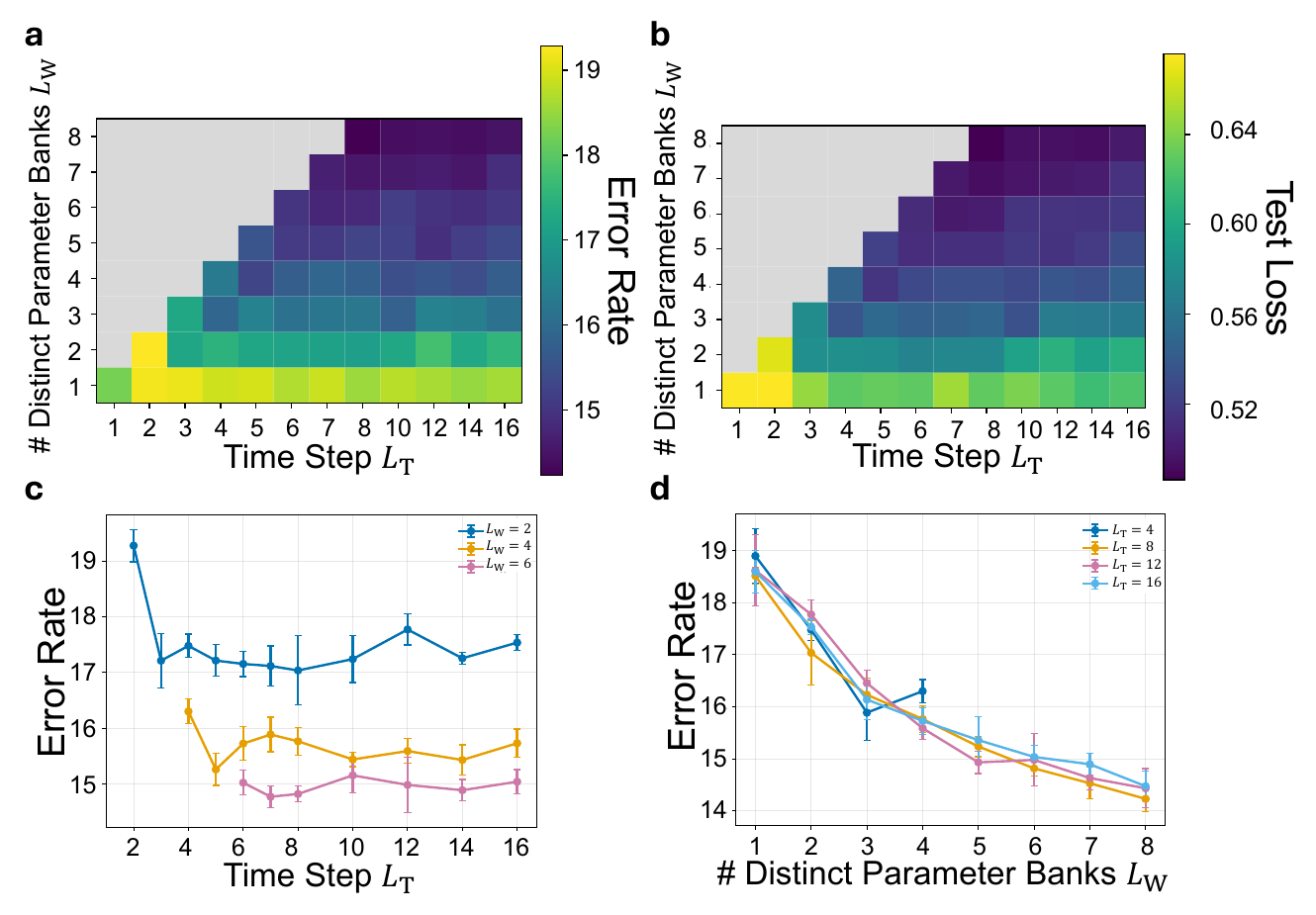}
\caption{TIDAL-Net performance heatmap on the SVHN classification when the width of hidden layers is fixed. \textbf{a}: Error. \textbf{b}: Loss.
The number of distinct layers $L_{\mathrm{W}}$ is varied along the y-axis, while the number of time steps $L_{\mathrm{T}}$ is varied along the x-axis.
By definition, $L_{\mathrm{T}}<L_{\mathrm{W}}$ is not a valid setting, and gray cells indicate such impossible configurations.
\textbf{c}: Extracted results from \textbf{a}. 
The number of time steps $L_{\mathrm{T}}$ is varied along the x-axis while keeping the number of distinct parameter banks $L_{\mathrm{W}}$ fixed.
\textbf{d}: Also extracted results from \textbf{a}. 
the number of distinct parameter banks $L_{\mathrm{W}}$ is varied along the x-axis while keeping the number of time steps $L_{\mathrm{T}}$ fixed.
Generally speaking, large $L_{\mathrm{W}}$ results in better performance. 
The best performance for a fixed $L_{\mathrm{W}}$ is, in general, not the case when $L_{\mathrm{W}}=L_{\mathrm{T}}$.
This means that repetitive usage can enhance the performance. 
}
\label{fig:fixlayer}
\end{figure}

Figure~\ref{fig:fixlayer}{$\mathbf{a}$} and {$\mathbf{b}$} show 2D color plots of the error rate and test loss, when varying $L_{\mathrm{T}}$ and $L_{\mathrm{W}}$. 
The color scale indicates the corresponding error rate or test loss, with lower values indicating better performance.
Change the number of total layers $L_{\mathrm{T}}$, i.e., changing how often layers are reused, does not change the total parameter count. 
Meanwhile, adding unique layers ($L_{\mathrm{W}}$) increases the number of parameters.
The diagonal of $L_{\mathrm{W}}=L_{\mathrm{T}}$ represents the case of an MLP-like resnet with no layers reused.
Conversely, the bottom row $L_{\mathrm{W}}=1$ corresponds to a single layer being reused multiple times, i.e., a stateless RNN.

Figure~\ref{fig:fixlayer}{$\mathbf{a}$} and {$\mathbf{b}$} exhibit common trends such as that more $L_{\mathrm{W}}$ results in better performance.
Importantly, within each row of $L_{\mathrm{W}} = \rm{const}$, the highest performance usually does not coincide with $L_{\mathrm{W}}=L_{\mathrm{T}}$ (diagonal of Fig.~\ref{fig:fixlayer}{$\mathbf{a}$} and {$\mathbf{b}$}), showing that repetition is beneficial.  

Figure~\ref{fig:fixlayer}{$\mathbf{c}$} isolates this effect more clearly, by showing line scans across $L_{\mathrm{T}}$ for various $L_{\mathrm{W}}$, with the effect strongest for networks with few distinct layers. 
It is worth re-emphasizing that changing $L_{\rm{T}}$ does not change the number of trainable parameters - the performance gains are purely thanks to repeating weights. 

TIDAL-Net combines the characteristics of RNNs and DNNs. 
In Fig.~\ref{fig:fixlayer}{$\mathbf{d}$}, by increasing the number of distinct parameter banks $L_{\rm{W}}$, TIDAL-Net can approach the high performance of DNNs.
However, when increasing the number of layers $L_{\rm{T}}$, we also notice RNN-like characteristics.
This includes the tendency for the performance to worsen eventually, if too many repetitions are used \cite{bengio1994learning}.
In any specific PNN, it is therefore key for designing a high-performing TIDAL-Net to balance between DNN-like and RNN-like characteristics.

\subsection*{Physical implementation requirements and timing trade-offs}
The numerical results above establish that temporal reuse can improve the computational utility of a finite set of parameter banks.
TIDAL-Net targets a scaling bottleneck complementary to increasing the size of an individual physical matrix.
Once a useful physical layer size is available, TIDAL-Net increases effective depth while reducing the number of independently implemented parameter banks required for that depth.
See also the fixed-parameter-budget experiments in Supplemental Information Section S2.


To clearly illustrate under which conditions TIDAL-Net confers a speed advantage, we analyze the interplay of the main timescales in the following:
Let $T_{\mathrm{upd}}$ denote the end-to-end time required to reprogram a physical weight bank.
Depending on the implementation, $T_{\mathrm{upd}}$ may include transfer of weight values from external memory, electronic control or digital-to-analog conversion, physical programming, and stabilization or settling of the device.
Let $T_{\mathrm{swt}}$ accordingly denote the time required to select or route among parameter banks that have already been configured.
TIDAL-Net is intended for the regime
\begin{align}
T_{\mathrm{swt}}& \ll T_{\mathrm{upd}},
\end{align}
so that chaining distinct layer-wise transformations does not require rewriting the physical weights at every logical layer.
For an MZI-based implementation, representative values discussed in Supplemental Information Section S10 are approximately $\mu\mathrm{s}$--$10\,\mu\mathrm{s}$ or longer for device-level weight tuning/programming, sub-nanoseconds to microseconds for external optical or electrical switching, picoseconds for the matrix--vector operation, and picoseconds to approximately $10\,\mathrm{ns}$ for the subsequent state-update block.
These values are device- and implementation-dependent, but illustrate the timescale separation targeted by TIDAL-Net.
For the $L_\mathrm{T}=12$ benchmarks considered here, even a device-programming contribution of order $1$--$10\,\mathrm{\mu s}$ per step would already correspond to order $10$--$100\,\mathrm{\mu s}$ over one forward pass, before additional memory-transfer, control, and state-update overheads. 

\begin{figure}[h]
\centering
\includegraphics[width=\textwidth]{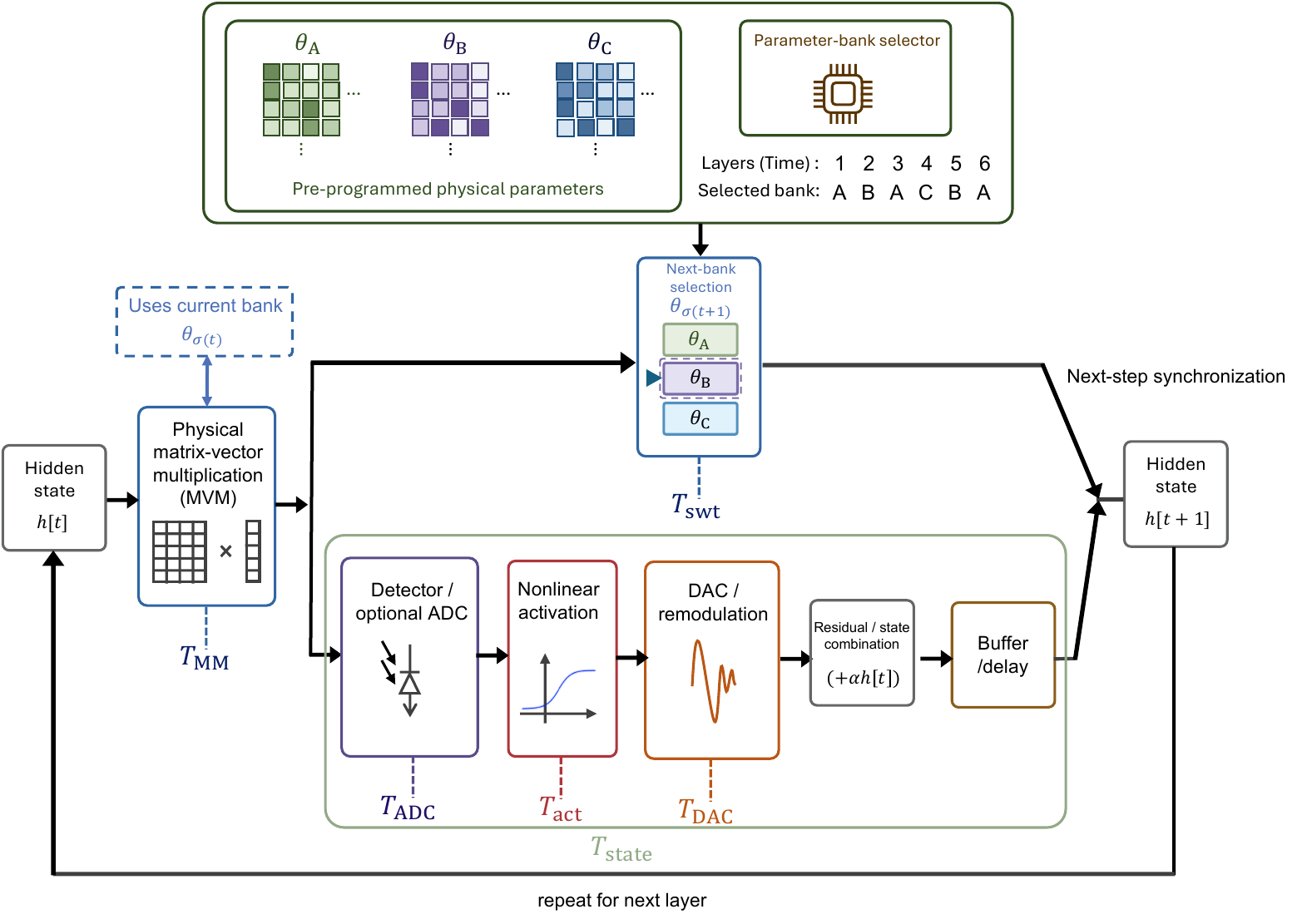}
\caption{Physical inference pipeline for TIDAL-Net.
A finite set of $L_\mathrm{W}$ physical parameter banks $\theta_k=\{W_\mathrm{hh}^{(k)},b_\mathrm{h}^{(k)}\}$ is programmed before inference and reused across logical layers.
At time step $t$, the current bank $\theta_{\sigma(t)}$ is used for the physical matrix--vector multiplication, with latency $T_{\mathrm{MM}}$.
After the matrix--vector multiplication, selection or preparation of the parameter bank for the next step, $\theta_{\sigma(t+1)}$, proceeds in parallel with the remaining state-update operations.
The state-update block includes detection and optional analog-to-digital conversion, nonlinear activation, signal remodulation, residual/state combination, and buffering or delay, with a total latency $T_{\mathrm{state}}$.
The next logical layer can start once both the next parameter bank and the updated hidden state are ready.
Under this coarse-grained scheduling model, the per-step latency is therefore
$T_{\mathrm{step}}=T_{\mathrm{MM}}+\max(T_{\mathrm{swt}},T_{\mathrm{state}})$, where $T_{\mathrm{swt}}$ denotes the parameter-bank switching or selection time.
ADC/DAC stages are implementation dependent and may be omitted in fully analog implementations.
The input projection and other auxiliary components are omitted from the schematic for clarity.}
\label{fig:physical_inference_pipeline}
\end{figure}

Figure~\ref{fig:physical_inference_pipeline} illustrates the corresponding target inference architecture.
At time step $t$, the currently selected parameter bank $\theta_{\sigma(t)}$ performs the matrix--vector multiplication.
After its output becomes available, selection or preparation of the parameter bank for the next logical layer can proceed in parallel with the remaining state-update operations.
Let $T_{\mathrm{MM}}$ denote the latency of the physical matrix--vector multiplication, and let $T_{\mathrm{state}}$ denote the subsequent state-update latency, including detection, nonlinear activation, signal regeneration or remodulation, residual/state combination, and buffering.
For implementations requiring them, analog-to-digital and digital-to-analog conversion are also included in this state-update path.
Under this coarse-grained scheduling model, a fully time-variable network that reprograms its physical parameters at every step has an approximate per-step latency
\begin{align}
T_{\mathrm{step}}^{\mathrm{time\mbox{-}variable}}
&=
T_{\mathrm{MM}}
+
\max
\left(
T_{\mathrm{upd}},
T_{\mathrm{state}}
\right),
\end{align}
whereas TIDAL-Net gives
\begin{align}
T_{\mathrm{step}}^{\mathrm{TIDAL}}
&=
T_{\mathrm{MM}}
+
\max
\left(
T_{\mathrm{swt}},
T_{\mathrm{state}}
\right).
\end{align}
Temporal reuse therefore does not remove the sequential forward operations associated with the effective depth $L_\mathrm{T}$.
Instead, it avoids the additional per-layer latency of complete physical weight reconfiguration when bank selection is substantially faster than reprogramming.
The timing advantage becomes small when $T_{\mathrm{state}}$ dominates both $T_{\mathrm{swt}}$ and $T_{\mathrm{upd}}$, or when the target platform does not support fast selection among preconfigured operators.
Likewise, TIDAL-Net does not remove conversion or nonlinear-processing costs; these costs must be included explicitly when evaluating a particular physical implementation.

The practical benefit of this reuse depends strongly on the physical resources and timescales of the target PNN platform.
Rather than identifying a single device response as the implementation timescale, Table~\ref{tab:physical_platform_comparison} shows several current PNN platforms and representative timescales for the various critical steps: weight programming as a proxy of $T_{\mathrm{upd}}$, bank switching $T_{\mathrm{swt}}$, matrix--vector propagation $T_{\mathrm{MM}}$, and nonlinear/state-update processing $T_{\mathrm{state}}$.
The end-to-end timescale is set by the critical path through these operations and therefore differs across platforms.
For the $64\times64$ recurrent banks used in the SVHN benchmark,
MZI-based photonic processors provide the closest match among the platforms considered here: the required bank dimension has already been demonstrated.
Here, matrix-vector multiplication is much faster than thermo-optic parameter tuning.
TIDAL-Net could, for the same depth, provide an inference cycle that is orders of magnitudes faster than a traditional time-multiplexing as a time-variable RNN. 
Such a TIDAL-Net realization would require multiple persistent banks connected by fast routing, making multi-bank integration the main remaining system-level step for this platform.
Table~\ref{tab:physical_platform_comparison} also shows that, while SLMs and DMDs offer substantially larger arrays, they suffer exactly from the slow reconfiguration speeds TIDAL-net is aiming to avoid;
SLM updates are millisecond-scale, while even DMDs that use binary weights and still have microsecond-scale update/readout times.
Nonvolatile photonic weights are attractive for persistent banks, but the demonstrated matrix scale considered here is smaller.

\begin{table*}[thbp]

\caption{\textbf{| Comparison of representative physical platforms for implementing reusable parameter banks.}
The listed quantities correspond to the main operations in the inference schedule of
Fig.~\ref{fig:physical_inference_pipeline}, including physical bank scale and representation, programming/tuning, bank selection, Matrix-Vector-Multiplication (MVM)/propagation, and conversion/nonlinear-processing overhead.
The values are drawn from representative experimental demonstrations and are used to identify characteristic platform trade-offs rather than to specify a single integrated system.
Benchmark-specific mapping is discussed in the main text, with further timescale
details in Supplemental Information Section S10.
}
\label{tab:physical_platform_comparison}

\centering
\small
\setlength{\tabcolsep}{3pt}

\begin{tabular}{
p{0.11\textwidth}
p{0.18\textwidth}
p{0.16\textwidth}
p{0.13\textwidth}
p{0.14\textwidth}
p{0.18\textwidth}
}

\hline

Platform
& Weight-bank scale / representation
& Weight update time $T_{\mathrm{upd}}$
& Bank switching $T_{\mathrm{swt}}$
& MVM / propagation $T_{\mathrm{MM}}$
& State overhead $T_{\mathrm{state}}$
\\

\hline

MZI mesh
& $4\times4$ academic ONN;
$64\times64$ industrial prototype;
continuous phase / amplitude
\cite{shen2017deep,ramey2020silicon}
& $\sim 4.7\,\mathrm{\mu s}$ thermal tuning time constant for a representative thermo-optic element
\cite{duan2021low}
& External switching / routing among preconfigured banks
& $\sim 200\,\mathrm{ps}$ MVM
\cite{ramey2020silicon}
& OEO conversion (PD + MRM):
$1.49\,\mathrm{GHz}$,
up to $4\,\mathrm{Gb/s}$;
ADC/DAC not required in this implementation
\cite{arahata2026silicon}
\\

\hline

Microring resonator array
& 16 tunable MRR weights in a 4-node chip;
analog wavelength-dependent weighting
\cite{tait2017neuromorphic}
& $\sim 4\,\mathrm{\mu s}$ thermal tuning in a representative microring-microheater implementation
\cite{atabaki2010optimization}
& External optical / electrical bank selection
& $\sim 48\,\mathrm{ps}$ optical-path delay;
GHz-class signal processing
\cite{tait2017neuromorphic}
& OEO conversion (Ge PD + MRM):
$1.49\,\mathrm{GHz}$,
up to $4\,\mathrm{Gb/s}$;
ADC/DAC not required in this implementation
\cite{arahata2026silicon}
\\

\hline

Spatial light modulator (SLM)
& $\sim 2\times10^{5}$ trainable parameters in a reconfigurable free-space ONN;
analog phase / amplitude
\cite{xia2023hardware}
& $\sim 17\,\mathrm{ms}$ update of the modulation pattern encoding the weights
(60\,Hz LCOS-SLM)
\cite{xue2024fully}
& No separate selection; operator change requires pattern update
& $\sim 1.3\,\mathrm{ns}$ optical propagation
\cite{xue2024fully}
& $42\,\mathrm{Hz}$ CMOS detection
($\sim 24\,\mathrm{ms/frame}$),
8-bit readout
\cite{xue2024fully}
\\

\hline

Digital micromirror device (DMD)
& $\sim 1.31\times10^{6}$ binary elements
($1280\times1024$ DMD);
binary amplitude
\cite{hornbeck1997digital}
& Up to $32\,\mathrm{kHz}$
($\sim 31\,\mathrm{\mu s}$ per binary modulation pattern)
\cite{ayoub2021high}
& No separate selection; operator change requires pattern update
& $\sim 2\, \mathrm{ns}$
free-space propagation in a representative DMD-based system
\cite{blanche2014crossconnect}
& InGaAs photodiode + TIA +
12-bit ADC
(up to $2\,\mathrm{MSample/s}$)
\cite{cox2019high}
\\

\hline

Nonvolatile photonic weights
& $16\times16$ fabricated PCM matrix;
analog or multilevel nonvolatile weights
\cite{feldmann2021parallel}
& $\sim 282\,\mathrm{ns}$ write;
$\sim 556\,\mathrm{ns}$ erase
for an electrically programmed GST cell
\cite{zhou2023memory}
& Passive WDM / external routing
& $>14\,\mathrm{GHz}$ photonic-compute bandwidth;
passive WDM routing
\cite{feldmann2021parallel,zhou2023memory}
& Photodetection +
$1.25\,\mathrm{MSample/s}$ DAQ;
ReLU implemented digitally
\cite{zhou2023memory}
\\

\hline

\end{tabular}
\label{table:timescale}

\end{table*}

Notably for SVHN, each reusable recurrent weight matrix has dimension $64\times64$ and, including its 64-dimensional bias, each parameter bank contains $64^2+64=4,160$ trainable hidden-layer parameters.
Thus, for example, the hidden-layer footprint of $33,280$ parameters at $L_\mathrm{W}=8$ corresponds to eight distinct $64\times64$ recurrent banks, rather than to one physical matrix containing more than $33,000$ independently addressed weights.
At the largest value $L_\mathrm{W}=12$ used in this comparison, the corresponding recurrent footprint is $49,920$ parameters distributed over twelve $64\times64$ banks.
The dimension of an individual recurrent bank is therefore already within the demonstrated $64\times64$ photonic scale discussed above.

The complete SVHN model also contains an input projection of dimension $64\times3072$ and an output projection of dimension $10\times64$.
The input projection therefore cannot be mapped directly onto a single $64\times64$ matrix processor.
For example, a straightforward block decomposition using $64\times64$ processors would divide the input transformation into 48 partial matrix--vector products whose outputs must subsequently be accumulated.
Alternatively, the input transformation could be implemented using preprocessing, a separate electronic or optical frontend, or a platform supporting a larger rectangular transformation.
This input stage is common to all values of $L_\mathrm{W}$ and is therefore not the physical resource whose duplication and temporal reuse are varied in our experiments, but it remains an important requirement for an end-to-end implementation.

For the Shakespeare next-token-prediction task, each reusable recurrent matrix in our numerical simulations has dimension $128\times128$, with $16,512$ trainable parameters per bank including the bias.
This recurrent matrix is larger than the representative $64\times64$ MZI prototype listed in Table~\ref{tab:physical_platform_comparison}.
An implementation using processors of that scale would therefore require a tiled realization---for example, four $64\times64$ matrix blocks for a straightforward $2\times2$ decomposition of each recurrent matrix---together with the associated routing and accumulation, or alternatively a larger future matrix processor.
The input and output transformations, of dimensions $128\times36$ and $42\times128$, would likewise require an appropriate rectangular mapping or tiling.
These are the matrix dimensions used in our benchmarks, not estimates of the minimum viable matrix size; we did not perform a systematic minimum-width search.


\section*{Discussion}\label{section_discussion_conclusion}

We introduced TIDAL-Net as an intermediate architecture between a stateless recurrent network and a fully untied deep network, motivated by physical neural networks in which forward dynamics can operate substantially faster than complete reconfiguration of the trainable physical parameters.
Rather than assigning an independently programmed parameter bank to every effective layer, TIDAL-Net reuses a finite set of banks over multiple time steps.
This separates the effective depth $L_\mathrm{T}$ from the number of distinct physical parameter banks $L_\mathrm{W}$, providing an additional architectural degree of freedom for trading physical parameter footprint against temporal reuse.

Across the image-classification and next-token-prediction benchmarks, increasing $L_\mathrm{W}$ from the $L_\mathrm{W}=1$ stateless-RNN limit consistently improved performance.
The experiments varying both $L_\mathrm{T}$ and $L_\mathrm{W}$ further show that the improvement cannot be attributed only to increasing the number of trainable parameters: for fixed $L_\mathrm{W}$, repeated use of the available banks can improve performance without increasing the hidden-layer parameter footprint.
At the same time, increasing effective depth indefinitely is not beneficial, and the preferred combination of $L_\mathrm{T}$ and $L_\mathrm{W}$ depends on the task.
The fixed-parameter-budget experiments in Supplemental Information Section S2 provide a complementary result: reducing the width of the individual banks merely to increase their number is not generally advantageous.
Temporal reuse should therefore be viewed as a way to obtain more computation from a physically feasible layer size, rather than as a universal substitute for increasing the size of the individual physical transformation.

More broadly, TIDAL-Net connects physical time-multiplexing with the established ideas of weight sharing and weight tying in digital neural networks.
Parameter sharing is widely used to impose structure and improve parameter efficiency in conventional architectures, including recurrent networks and Transformers~\cite{takase2023lessons,bae2025relaxed}.
Here, the same principle is considered from a physical-resource perspective: reusing an already implemented transformation may be considerably easier than either duplicating it or repeatedly transferring and rewriting its parameters.
The periodic switching scheme studied here is only one member of a broader class of partially time-multiplexed architectures.
Supplemental Information Section S1 shows that non-periodic switching schedules can provide additional, although modest, improvements.
A natural extension is therefore to optimize the switching schedule jointly with the parameter banks under the constraints of a specific physical platform.

Several limitations remain.
The present work evaluates TIDAL-Net numerically rather than through a hardware prototype, and the component-level values summarized in
Table~\ref{tab:physical_platform_comparison} are drawn from different experimental demonstrations rather than from one integrated system.
The benchmark architectures were also not optimized to determine the minimum physical matrix dimensions required for each task.
Moreover, practical implementations will introduce additional routing, synchronization, control, calibration, signal-conversion, and device-nonideality constraints.
Finally, we do not provide an end-to-end energy model: optical nonlinearities may require substantial pump power, whereas optoelectronic implementations incur conversion and processing costs, so fewer parameter banks do not by themselves imply lower energy consumption.

An important next step is an experimental realization combining persistent physical weight banks, fast optical or electrical routing, and a reusable nonlinear state-update path. For the SVHN-scale recurrent core considered here, multiple $64\times64$ parameter banks provide a concrete experimental target. 
Such a system would directly test whether the separation between slow parameter programming and fast forward dynamics can be exploited through architectural reuse. More broadly, our results suggest that physical constraints can motivate neural-network architectures matched to the resources and timescales of the underlying hardware.

\section*{Methods}

\subsection*{TIDAL-Net}

TIDAL-Net is a temporally unrolled neural-network architecture designed to reuse a finite number of physical weight banks over multiple effective layers. 
The hidden state at time step $t$ is denoted by $h[t] \in \mathbb{R}^{H}$, where $H$ is the hidden-state dimension. 
For an input $x[t]$, the hidden state evolves according to
\begin{equation}
h[t+1]
=
\alpha h[t]
+
f\left(
W_\mathrm{xh}x[t] + W_\mathrm{hh}[t]h[t] + b_\mathrm{h}[t]
\right),
\label{eq:tidal_update}
\end{equation}
where $W_{\mathrm{xh}}$ is the input-to-hidden weight matrix, $W_{\mathrm{hh}}[t]$ is the hidden-to-hidden weight matrix used at time step $t$, $b_h[t]$ is the corresponding hidden bias, $f$ is the nonlinear activation function, and $\alpha$ is a trainable residual coefficient. 
The residual term $\alpha h[t]$ acts as a skip-like connection between consecutive effective layers and was included to improve training stability.

Using column-vector notation, the input-to-hidden, hidden-to-hidden, and hidden-to-output matrices have dimensions $W_{\mathrm{xh}}\in\mathbb{R}^{H\times X}$, $W_{\mathrm{hh}}^{(k)}\in\mathbb{R}^{H\times H}$, and $W_{\mathrm{hy}}\in\mathbb{R}^{Y\times H}$, respectively, where $X$, $H$, and $Y$ denote the input, hidden-state, and output dimensions.
Each switched hidden-layer bank $\theta^{(k)}=\{W_{\mathrm{hh}}^{(k)},b_{\mathrm{h}}^{(k)}\}$ therefore contains one $H\times H$ recurrent matrix and one $H$-dimensional bias vector.

The central feature of TIDAL-Net is that $W_{\mathrm{hh}}[t]$ and $b_{\mathrm{h}}[t]$ are not independently assigned for every time step. 
Instead, the network uses a finite set of $L_{\mathrm{W}}$ distinct hidden-layer parameter banks,
\begin{equation}
\theta^{k} = \{W_{\mathrm{hh}}^{k}, b_{\mathrm{h}}^{k}\}, 
\qquad
k = 0, \ldots, L_\mathrm{W}-1,
\end{equation}
and switches among them over $L_\mathrm{T}$ temporal steps. 
In the simplest periodic switching scheme used in this work, the parameter bank at time step \(t\) is selected as
\begin{equation}
\theta[t] = \theta^{t \bmod L_\mathrm{W}}.
\label{eq:periodic_switching}
\end{equation}
Thus, $L_\mathrm{T}$ denotes the effective network depth, whereas $L_\mathrm{W}$ denotes the number of distinct hidden-layer parameter banks. 
The case $L_\mathrm{W} = 1$ corresponds to a stateless recurrent neural network that repeatedly applies the same hidden transformation. 
The case $L_\mathrm{W} = L_\mathrm{T}$ corresponds to an untied multilayer network in which each effective layer has its own hidden parameters. 
The intermediate regime $1 < L_\mathrm{W} < L_\mathrm{T}$ is the main TIDAL-Net regime considered in this work.

The final hidden state $h[L_\mathrm{T}]$ is mapped to the prediction by
\begin{equation}
\hat{y} = W_\mathrm{yh}h[L_\mathrm{T}] + b_\mathrm{y},
\label{eq:output}
\end{equation}
where $W_\mathrm{hy}$ and $b_\mathrm{y}$ are the hidden-to-output weight matrix and output bias, respectively. 
Unless otherwise stated, the nonlinear activation function was ReLU.

\subsection*{Training and optimization}

All models were trained by unrolling the temporal dynamics over $L_\mathrm{T}$ steps and applying backpropagation through time. 
The trainable parameters were the input-to-hidden weights, the output weights, the residual coefficient $\alpha$, and the finite set of hidden-layer weight banks and biases. 
The periodic reuse of hidden-layer parameters was handled directly in the computational graph, so that gradients from all time steps using the same parameter bank were accumulated during backpropagation.

For a mini-batch of $N_{\mathrm{batch}}$ samples, the loss was computed as the average cross-entropy loss over the mini-batch:
\begin{align}
\mathcal{L}_{\mathrm{batch}}
&=
\frac{1}{N_{\mathrm{batch}}}
\sum_{k=1}^{N_{\mathrm{batch}}}
\ell(y_{\mathrm{target}}^{k}, \hat{y}^{k}),
\end{align}
where $\ell$ denotes the cross-entropy loss. 
For classification tasks, the target labels were represented as class indices, and the model output was interpreted as unnormalized logits. 
For next-token prediction, the output logits represented the probability distribution over the token vocabulary.

Parameters were optimized using Adam. 
Unless otherwise specified, we used a learning rate of $0.001$ with the standard Adam parameters $\beta_1 = 0.9$, $\beta_2 = 0.999$, and $\epsilon = 10^{-8}$. 
The hidden-layer, input-layer, and output-layer weights were initialized using Kaiming initialization for ReLU nonlinearities. 
Biases were initialized to zero. 
The residual coefficient $\alpha$ was initialized to $1.0$ and optimized during training.

For model selection and hyperparameter tuning, the original training set was split into training and validation subsets with a $9:1$ ratio. 
The validation set was used to check performance during training and to select model settings such as the number of hidden units and the switching parameters $L_\mathrm{T}$ and $L_\mathrm{W}$. 
The final reported results were evaluated on the test set.

\subsection*{SVHN image-classification task}

We evaluated TIDAL-Net on the Street View House Numbers (SVHN) dataset \cite{svhn}. 
The task is a ten-class classification problem, where each input is a $32 \times 32$ RGB image containing a central digit. 
We used the standard SVHN training and test splits, consisting of 73,257 training images and $26,032$ test images.

For the SVHN experiments, each image was converted to a tensor and reshaped into a 3072-dimensional vector. 
During training, data augmentation was applied to improve generalization. 
Each image was padded by $4$ pixels on each side and randomly cropped back to $32 \times 32$ pixels. 
We also applied random color jitter, perturbing brightness, contrast, and saturation by up to $10\%$ and hue by up to $5\%$. 
Pixel-wise Gaussian noise was added after conversion to floating-point values. 
The images were then normalized channel-wise and flattened before being used as input to the network.

Each $32\times32$ RGB image was flattened into an input vector with dimension $X=3072$. 
Unless otherwise specified, we used a hidden-state dimension of $H=64$ and an output dimension of $Y=10$. Thus, the input-to-hidden matrix had size
$W_{\mathrm{xh}}\in\mathbb{R}^{64\times3072}$, each recurrent weight bank had size $W_{\mathrm{hh}}^{(k)}\in\mathbb{R}^{64\times64}$, and the hidden-to-output matrix had size $W_{\mathrm{hy}}\in\mathbb{R}^{10\times64}$.
Including its $64$-dimensional bias vector, each recurrent bank contained $64^2+64=4{,}160$ trainable hidden-layer parameters.

For image classification, the same image vector was injected at every time step,
\begin{align}
x[t] = x_{\mathrm{image}},
\qquad
t = 0, \ldots, L_\mathrm{T}-1.
\end{align}
The final hidden state after $L_\mathrm{T}$ steps was used for classification. 
Unless otherwise specified, the hidden dimension was $H = 64$, and the models were trained for 900 epochs. 
The performance was reported as the test error rate and the cross-entropy test loss.

\subsection*{Next-token prediction task}

We also evaluated TIDAL-Net on a character-level next-token prediction task using Shakespearean \cite{folger_shakespeare} text. 
The corpus contained 9,044,939 characters. 
The text was preprocessed by converting uppercase letters to lowercase and mapping variant characters to a common representation. 
Infrequent characters with an occurrence rate below $0.01\%$ were replaced by a placeholder token. 
The final vocabulary contained 42 tokens.

The tokenized text was split into training and test subsets using a contiguous $90:10$ split. 
The integer token sequence was then divided into fixed-length samples. 
For each sample, the first $T$ tokens were used as input, and the following token was used as the target. 
Token indices were embedded into continuous input vectors before being fed into the network. 
In contrast to the SVHN task, the input at each time step was a different token embedding, so that $x[t]$ represented the token at position $t$ in the input sequence.

Each input token was represented by a $36$-dimensional embedding.
For the next-token prediction experiments, we used a hidden-state dimension of $H=128$ and a vocabulary and output dimension of $V=Y=42$.
The embedding table therefore had size $E\in\mathbb{R}^{42\times36}$, the input-to-hidden matrix had size $W_{\mathrm{xh}}\in\mathbb{R}^{128\times36}$, each recurrent weight bank had size $W_{\mathrm{hh}}^{(k)}\in\mathbb{R}^{128\times128}$, and the hidden-to-output matrix had size $W_{\mathrm{hy}}\in\mathbb{R}^{42\times128}$.
Including its $128$-dimensional bias vector, each recurrent bank contained $128^2+128=16{,}512$ trainable hidden-layer parameters.

In the numerical experiments, the models were trained for $200$ epochs. 
To reduce overfitting in the recurrent setting, we used variational dropout. 
In this dropout scheme, dropout masks were sampled once per mini-batch and shared across time steps, rather than independently resampled at every time step. 
Separate masks were applied to the input-to-hidden, hidden-to-hidden, and hidden-to-output transformations. 
The model was evaluated using the test error rate and cross-entropy test loss.

\subsection*{Hidden layers parameter counting}

The parameter footprint reported in the main figures refers primarily to the number of trainable hidden-layer parameters associated with the physical weight banks. 
For a hidden dimension $H$ and $L_\mathrm{W}$ distinct hidden-layer parameter banks, the hidden-layer parameter count is
\begin{align}
N_{\mathrm{hidden}}
=
L_\mathrm{W}(H^2 + H),
\label{eq:hidden_param_count}
\end{align}
where $H^2$ counts the entries of the hidden-to-hidden matrix and $H$ counts the hidden bias. 
Unless otherwise specified, this footprint excludes the input-to-hidden weights, output weights, and residual coefficient. 
This convention was used because the hidden-layer weight banks are the primary physical resources whose duplication or reuse is investigated in this work.

For experiments in which $L_\mathrm{W}$ was varied while $H$ and $L_\mathrm{T}$ were fixed, increasing $L_\mathrm{W}$ increased the hidden-layer parameter footprint approximately linearly. 
For fixed-budget experiments, we instead adjusted the hidden dimension $H$ as $L_\mathrm{W}$ was changed so that $L_\mathrm{W}(H^2 + H)$ remained approximately constant. 
Because $H$ must be an integer, the exact number of parameters varied slightly across $L_\mathrm{W}$ values. 
The hidden dimensions used in the fixed-budget experiments are reported in the corresponding table in Supplemental Information Section S2.

We emphasize that this parameter count is not intended to provide an architecture-independent measure of computational expressivity. 
In physical neural networks, the relation between trainable parameters, physical degrees of freedom, and expressivity depends on the physical substrate, architecture, and task. 
The parameter counts used here should therefore be interpreted as the number of explicitly trained or controllable hidden-layer degrees of freedom in the simulated TIDAL-Net model.

\subsection*{Numerical experiments}

We performed two main types of numerical experiments. 
First, we fixed the effective depth $L_\mathrm{T}$ and varied the number of distinct hidden-layer parameter banks $L_\mathrm{W}$. 
This experiment interpolates between the stateless RNN limit, $L_\mathrm{W} = 1$, and the untied multilayer limit, $L_\mathrm{W} = L_\mathrm{T}$. 
It was used to evaluate whether switching among multiple preconfigured weight banks improves performance over repeatedly applying a single fixed transformation. (Fig.~\ref{fig:svhn_results} and Fig.~\ref{fig6:NLP})

Second, we fixed the hidden-layer width $H$ and varied both $L_\mathrm{T}$ and $L_\mathrm{W}$ to examine the effect of temporal repetition and the transition from the RNN-like limit to the DNN-like limit. This experiment was used for Fig.~\ref{fig:fixlayer}. Fixed-hidden-parameter-budget experiments, in which $H$ was adjusted so that $L_\mathrm{W}(H^2 + H)$ remained approximately constant, are reported in Supplemental Information Section~S2.
This experiment tests whether it is preferable, under a fixed physical-parameter budget, to implement a small number of wide weight banks or a larger number of narrower weight banks. (Fig.~S7)

\subsection*{Physical implementation assumptions}

TIDAL-Net is motivated by physical neural-network platforms in which the trainable physical weights are slower to tune than the dynamical degrees of freedom used during inference. 
The model assumes that a small number of weight banks can be prepared or calibrated before inference, and that inference proceeds by fast switching among these preconfigured banks. 
Thus, the trainable weights do not need to be fully reconfigured at every inference step.

In a physical implementation, one TIDAL-Net step corresponds to a propagation, nonlinear activation, hidden-state update, and switching cycle. 
The total inference time is therefore approximately proportional to $L_\mathrm{T}$ times the duration of one such cycle. 
This introduces a trade-off between physical network size and temporal computation time. 
TIDAL-Net is expected to be most relevant in regimes where duplicating many physical layers or fully reconfiguring all weights at every inference step is more costly than switching among a small number of preconfigured weight banks.

Training in experiments could be performed using existing physical neural-network training strategies. 
One possible approach is to train a calibrated differentiable model of the physical system in silico and then program the obtained parameters into the device. 
Another approach is hardware-in-the-loop training, such as physics-aware training, where the physical system is used for forward measurements while parameter updates are computed using a differentiable surrogate model or experimentally estimated gradients. 
In both cases, the physical weights can be updated on the slower training or calibration timescale, while inference requires only fast switching among trained weight banks.

\subsection*{Statistics and reproducibility}

The reported curves were obtained from repeated training runs with different random seeds where applicable. 
Error bars represent the variability across runs. 
All numerical experiments were implemented in PyTorch. 
Random seeds were fixed for each run to improve reproducibility, and the same preprocessing and evaluation procedures were used for all models within each comparison.

\section*{Data availability}

Source Data are provided with this paper. The numerical source data underlying the graphs, charts, and tables in the main manuscript and Supplementary Information are included as labelled CSV files in the Source Data folder. Lightweight precomputed numerical outputs sufficient to regenerate the manuscript figures are included in the Code Ocean capsule provided for peer review and will be made publicly available with a permanent DOI/link upon acceptance or publication. The publicly available datasets analysed during the current study are available from their original sources: the SVHN dataset, the MNIST dataset, and the Shakespeare text corpus from the Folger Shakespeare Library.



\section*{Code availability}
Code has been provided to the editors and reviewers via Code Ocean for peer review; a public DOI/link will be provided upon acceptance/publication.

\bibliographystyle{unsrt}
\bibliography{arxiv-sn-biblliography}

\section*{Acknowledgements}
We thank Felix Köster for fruitful feedback and discussions. 
This study was supported in part by a Grant-in-Aid for Transformative Research Areas (A) (JP22H05197), a Grant-in-Aid for JSPS Fellows (JP24KJ0868), Grant-in-Aid for Exploratory Research (JP25K22227),  JST-ALCA-Next (JPMJAN25F1), JST FOREST Program (JPMJFR2448) and SECOM Science and Technology Foundation.

\clearpage

\begin{appendices}

\raggedbottom

\renewcommand{\thefigure}{S\arabic{figure}}
\renewcommand{\thetable}{S\arabic{table}}

\renewcommand{\figurename}{Supplementary Fig.}
\renewcommand{\tablename}{Supplementary Table}

\renewcommand{\thesection}{S\arabic{section}}
\renewcommand{\thesubsection}{\thesection.\arabic{subsection}}
\renewcommand{\thesubsubsection}{\thesection.\arabic{subsection}.\arabic{subsubsection}}

\renewcommand{\thefigure}{S\arabic{figure}}
\renewcommand{\thetable}{S\arabic{table}}
\renewcommand{\theequation}{S\arabic{equation}}

\renewcommand{\figurename}{Supplementary Fig.}
\renewcommand{\tablename}{Supplementary Table}

\section{Weight mixing: extension of periodic switching}
In the main manuscript, we focus on periodic reuse of the prepared parameters $\theta[t]=\{W_{\rm{hh}}[t], b_{\rm{h}}[t]\}$.
In other words, we use a distinct set of prepared hidden-layer parameters, $\theta_k=\{W_\mathrm{hh}^{(k)}, b_\mathrm{h}^{(k)}\}_{k=1}^{L_\mathrm{W}}$.
We note that the periodic switching of a layer architecture considered in the main text is not the only possible TIDAL-Net structure. 
For example, longer-range skip connections or branched architectures could also be implemented if additional physical memory, delay lines, or routing paths are available. 

Among various possible extensions, a natural extension of periodic switching is weight modulation, which designs how the prepared weights $(W_{\rm{hh}}^1,W_{\rm{hh}}^2,...,W_{\rm{hh}}^{\rm{L_{W}}})$ are combined at each time step.
Similar ideas have been explored in dynamic parameterized neural networks, such as CondConv~\cite{yang2019condconv} and dynamic convolution~\cite{chen2020dynamic}, where multiple pre-defined convolution kernels are combined using input-dependent routing or attention coefficients.
In contrast, the simple variant considered here uses time-indexed trainable coefficients rather than input-dependent routing coefficients.
Periodic switching can be viewed as a special case of this formulation, in which one coefficient is one, and all others are zero at each time step.
This also relies on fast inference and exploits the same timescale separation.

Using mixing coefficients $s_k[t]$, the recurrent weight matrix at time step $t$ is defined as
\begin{align}
    W_{\rm{hh}}[t]&=\sum_{k=1}^{L_{\mathrm{W}}}s_k[t]W_{\rm{hh}}^{k} \label{eq:mix_weight}.
\end{align}
In the implementation used here, the coefficients $s_k[t]$ are trained directly as real-valued parameters. They are initialized as normalized positive coefficients.
Here, we do not constrain them to remain non-negative or to sum to one during training, although such restrictions could be added, where physically required.
Instead of periodically repeating the parameters $\theta[t]$, this architecture allows the weight matrix term at each step to be determined through a mixture of the prepared components.
For simplicity, we apply mixing only to the recurrent weight matrices. The bias term is treated as an unconstrained time-dependent trainable vector $b_\mathrm{h}[t]$.
In this system, the trainable parameters are the weight set, mixing coefficients, and bias terms.

Under this architecture, we performed the same SVHN classification task setting as in the main manuscript. 
The training algorithm and task settings are the same in the case of periodic switching.
We initialize the coefficients $s_k[t]$ as $s_k^{(0)}[t]$, where
\begin{align} 
\pi[t]&=t\ \mod L_{\mathrm{W}}\\
s_k^{(0)}[t]&=\frac{\mu}{L_\mathrm{W}+\mu}\mathbf{1}[k=\pi[t]]+ \frac{1}{L_\mathrm{W}+\mu}. \label{eq:mix_initialize}
\end{align}
When $\mu=0$, this initialization gives the uniform mixture $s_k^{(0)}[t]=1/L_\mathrm{W}$.
As $\mu \rightarrow \infty$, it approaches hard periodic switching, where $s_{\pi[t]}^{(0)}[t]\rightarrow 1$ and the other coefficients approach zero. 
In our experiments, we set $\mu=1.0$, where coefficients are biased to, but not exclusively concentrated in, the periodic switching pattern.

The hidden-state update is therefore
\begin{align}
h[t+1]&=\alpha h[t]+\phi\left(\sum_{k=1}^{L_\mathrm{W}} s_k[t] W_{\mathrm{hh}}^{(k)} h[t]+b_\mathrm{h}[t]+W_{\mathrm{xh}}x[t]\right).
\end{align}

\begin{figure}[h]
\centering
\includegraphics[width=1.0\textwidth]{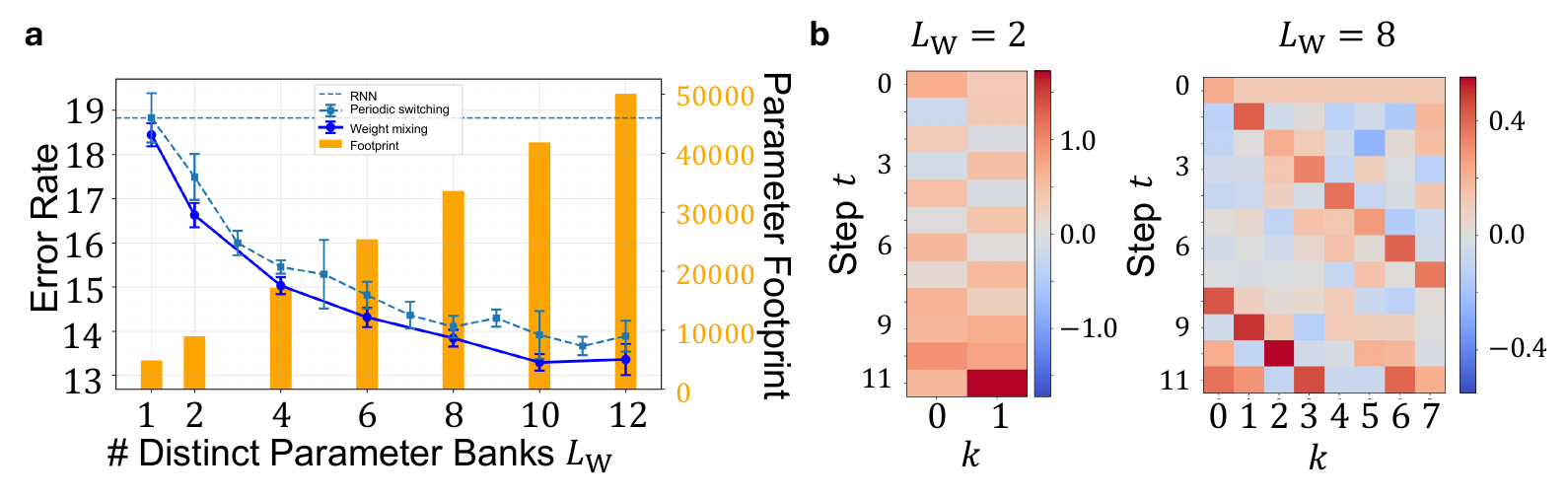}
\caption{\textbf{|Performance of weight mixing on TIDAL-Net.} $\mathbf{a}$: Error Rate on SVHN image classification tasks. Both the mixing model (solid blue plot) and the periodic switching model (dot blue plot) results are similar. Orange bars show the number of trainable parameters. $\mathbf{b}$: Weight mixing coefficient $s_k[t]$ values. The coefficient distributions loosely resemble periodic switching due to the seeding, but clearly includes additional mixing among weights. }\label{fig:mixing}
\end{figure}

Figure~\ref{fig:mixing} shows the performance of weight-mixing models.
The performance of mixing (dark blue) is similar to or slightly exceeding that of periodic switching (light blue) on the SVHN image classification task. 
This suggests that, in this one-shot SVHN setting, continuous weight mixing does not substantially improve over simple periodic switching, but it may yield further performance improvements at a negligible parameter cost.
Note that the error bars of both simulations are overlapping. 
Confirming whether the advantage of mixing exists in general, is an open question requiring more computational resources.
In Fig.~\ref{fig:mixing}~{$\mathbf{b}$}, you can see the weight coefficient $s_k[t]$ after training for two cases.
We can see the dominant structure along the main-diagonal (and wrap around), from the initial starting seed. 
However, we can also clearly see that off-diagonal elements vary and are not always close to zero. 
This shows that indeed the training leads to weights that are well mixed. 
For example, the first temporal layer $t=0$ in both cases is a strong mixture of all weight matrices.

We take these results as further indication that the space of time-multiplexed PNNs should be further explored, with a particular focus on the kinds of extensions that are cheap for each particular hardware platform, including weight mixing where possible.

\section{TIDAL-Net design under a fixed hidden-parameter budget}\label{appendix_parameters}
Considering the other results of the main manuscripts and Supplemental Information, one might be tempted to summarize it as the well-established fact that a deep neural network is, in many cases, much more powerful than a shallow one. 
In this sense, TIDAL-Net emphasizes the inherent power of using multiple distinct layers, instead of just a single one.
Therefore, one may wonder whether it is worth considering artificially increasing the set of distinct weights when building a PNN from various physical components. 
For example, one may be able to use the same lab space, or wafer space for an integrated chip, or temporal range for a time-multiplexed network, to either implement one large layer, or two smaller layers. 
Here, we want to consider such a scenario and investigate the performance using a fixed parameter budget.

We choose a size of roughly $33000$ trainable parameters, and divide these among $L_{\mathrm{W}}$ distinct weight matrices $W_{\mathrm{hh}}$, as shown in Table~\ref{tab_fix_param}. 
To isolate the effect we care for, we also use fixed random input parameters $W_{\mathrm{xh}}$, as otherwise the implicit number of trainable parameters of the input layer would vary with $L_{\mathrm{W}}$ as well.

\begin{figure}[h]
\centering
\includegraphics[width=0.9\textwidth]{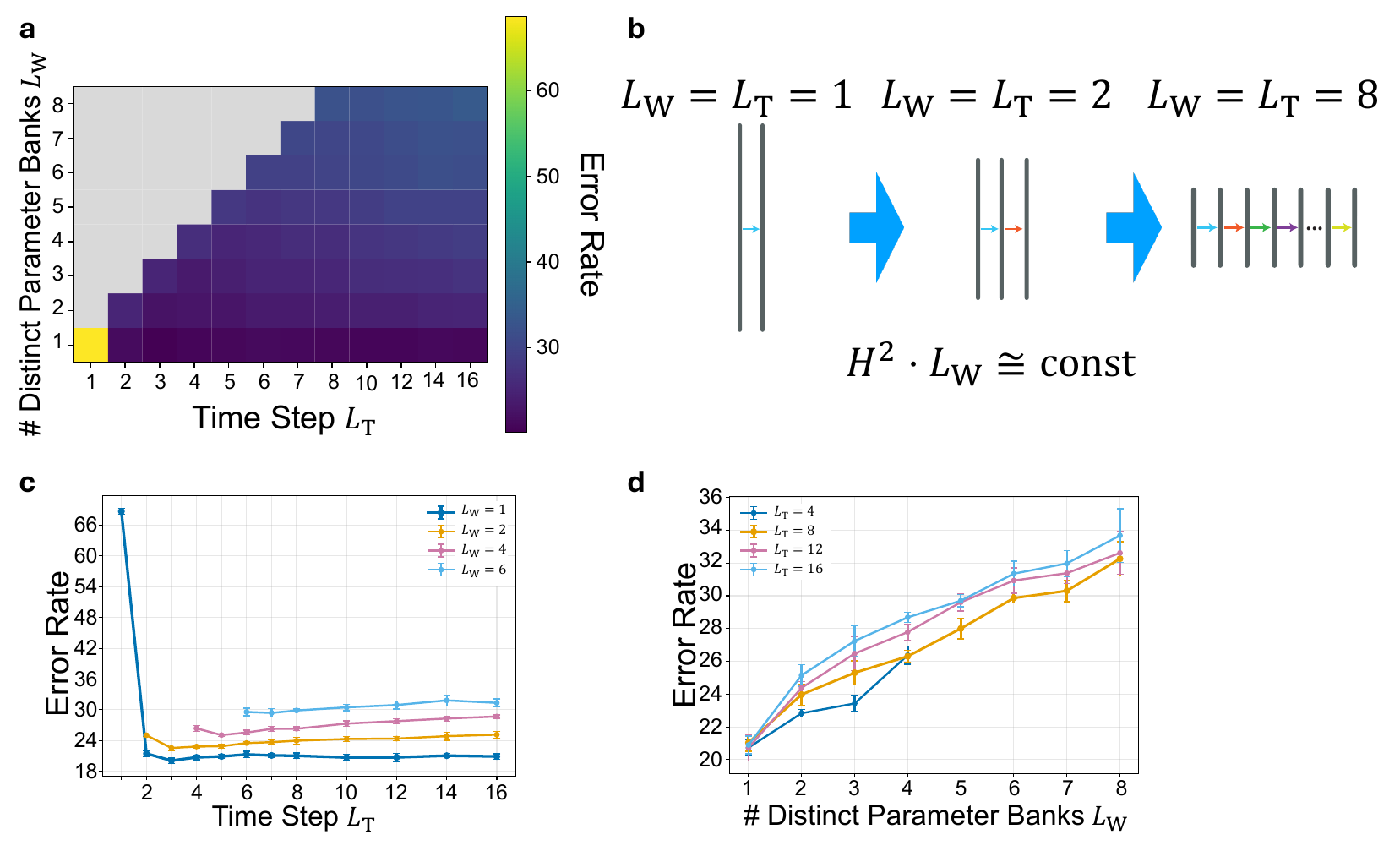}
\caption{
\textbf{TIDAL-Net performance on SVHN classification under a fixed hidden-parameter budget.}
\textbf{a}, Classification error when the number of distinct weight banks $L_\mathrm{W}$ and the number of temporal steps $L_\mathrm{T}$ are varied. Configurations with $L_\mathrm{W} > L_\mathrm{T}$ are not valid and are shown as gray cells.
\textbf{b}, Schematic examples of model structures on the diagonal $L_\mathrm{W}=L_\mathrm{T}$. Because the hidden-layer parameter budget is fixed, increasing $L_\mathrm{W}$ requires reducing the hidden dimension $H$ of each weight bank.
\textbf{c}, Classification error as a function of $L_\mathrm{T}$ for fixed $L_\mathrm{W}$. For a fixed hidden-parameter budget, a smaller $L_\mathrm{W}$ corresponds to wider hidden layers and generally gives better performance. The best performance is often obtained at $L_\mathrm{T} > L_\mathrm{W}$, indicating that temporal repetition can improve performance without increasing the number of trainable hidden-layer parameters.
\textbf{d}, Classification error as a function of $L_\mathrm{W}$ for fixed $L_\mathrm{T}$. Increasing $L_\mathrm{W}$ generally worsens performance because each weight bank becomes narrower under the fixed-budget constraint. Larger $L_\mathrm{T}$ values tend to perform better than smaller ones, further supporting the contribution of temporal repetition.
}\label{fig:Appendix_param_heatmap}
\end{figure}

Figure~\ref{fig:Appendix_param_heatmap}{\textbf{a}} shows a 2D plot of the performance of TIDAL-Net on the SVHN image classification task under this fixed parameter budget. 
Here, we vary the number of distinct parameter banks $L_{\mathrm{W}}$ on the y-axis and the number of layers $L_{\mathrm{T}}$ (temporal length) on the x-axis. 
Note that the parameter budget is independent of $L_{\mathrm{T}}$, as this merely represents repeating previous layers. 
When $L_{\mathrm{W}}$ changes, it also changes the size of the hidden layers as per Table~\ref{tab_fix_param} (see Fig.~\ref{fig:Appendix_param_heatmap}{\textbf{b}}). 
Cases of $L_{\mathrm{W}}>L_{\mathrm{T}}$ are impossible and represented by gray cells.
We use the previously presented periodic switching of weights.

Figure~\ref{fig:Appendix_param_heatmap}{\textbf{c}} shows the extracted result from Figure~\ref{fig:Appendix_param_heatmap}{\textbf{a}}, showing the cases when fixing the number of weights $L_{\mathrm{W}}$.
As can be seen, the performance is best with $L_{\mathrm{W}}=1$ and $L_{\mathrm{T}}>1$. 
This corresponds to the case of a single, large layer, with a few repetitions.

On the other hand, Fig.~\ref{fig:Appendix_param_heatmap}{\textbf{d}} shows the results when we vary the number of weights $L_{\mathrm{W}}$ while keeping the number of steps $L_{\mathrm{T}}$.
In contrast to Fig.~\ref{fig:fixlayer}\textbf{d}, the performance increases when $L_{\mathrm{W}}$ increases.
It is because the width of the layer decreases when the performance increases, and the total number of parameters is fixed.

These results indicate that, at least in the tested cases, it is not worth sacrificing the size of individual layers in order to increase the number of distinct layers in TIDAL-Net. 
TIDAL-Net is beneficial when physical constraints, such as size or heat dissipation limits, are already setting the maximum size of each trainable layer in terms of the dimension of $\theta^l$. 
But as Fig.~\ref{fig:Appendix_param_heatmap} shows, halving the size of each layer, in order to increase the number of layers, does not increase the performance under the conditions we tested.
Nevertheless, from the 2D-plots in Fig.~\ref{fig:Appendix_param_heatmap} we can see that for each fixed $L_{\mathrm{W}}$ row, the best performance is typically achieved for $L_{\mathrm{T}}>L_{\mathrm{W}}$. i.e., when at least some layers are repeated, showcasing once more one of the main motivations for TIDAL-Net.

\begin{table}[h]
\caption{The number of hidden nodes used in Fig.~\ref{fig:Appendix_param_heatmap}. 
Here, the trainable parameters consist of the network weights $W_{\rm{hh}}$ and bias terms $b_{\rm{h}}$.
This excludes the input-hidden network $W_{\rm{xh}}$, hidden-output network $W_{\rm{hy}}$ and residual parameters $\alpha$. Note: Total parameter counts fluctuate because the dimension of hidden nodes needs to be an integer.}\label{tab_fix_param}
\begin{tabular}{@{}lllllllll@{}}
\toprule
\# different layers $L_{\rm{W}}$   & 1 & 2  & 3 & 4 & 5 & 6 & 7 & 8 \\
\midrule
hidden nodes per layer $H$   & 192 & 128 & 104 & 91 & 80 & 73 & 68 & 64 \\
\# trainable parameters & 37057   & 32897  & 32761 & 33489 & 32401 & 32413 & 32845 & 33281\\
\botrule
\end{tabular}
\end{table}

\clearpage

\section{TIDAL-Net with an Oscillator Network}
As explained in the Introduction, spintronics is a promising approach to achieving energy-efficient PNNs.
In the PNN implementation with spintronics, nodes in hidden layers are something that oscillate, such as magnetic spins.
Oscillator networks introduce nonlinearity into computing in a different way from the standard activation functions like ReLU.
To investigate the generality regardless of the activation function, we conduct an experiment on TIDAL-Net where the nonlinear activations are achieved with oscillations.

Let the hidden states $h[t]$ evolve with the help auxiliary intermediate states $h_{\rm{c}}^{(\kappa)}\in \mathbb{R}^H, \kappa\in[0,...,\eta]$ evolve according to:
\begin{align}
h_{\rm{c}}^{(0)}&:=h[t-1]\\
h_{\rm{c}}^{(\kappa+1)}&=h_{\rm{c}}^{(\kappa)}+\Delta t\cdot\left( \left(\gamma_{\rm{p}}-\gamma_{\rm{nl}}\left(h_{\rm{c}}^{(\kappa)}\right)^2\right)h_{\rm{c}}^{(k)}+W_{\rm{xh}}x[t]+b_{\rm{h}}[t]\right)\\
h[t]&=\alpha h[t-1] + h_{\rm{c}}^{(\eta)},
\end{align}
where $\gamma_{\rm{p}}, \gamma_{\rm{nl}} \in \mathbb{R}$ are the trainable parameters of the spintronics system and $\Delta t\in\mathbb{R}_{>0}$ is the parameter for the integration over time in the numerical calculation.
In our experiment, $\Delta t$ is fixed such that $\eta$ is $3$.
This equation corresponds to the simplified Slavin's model \cite{Plouet2025} as follows:
\begin{align}
\frac{dx}{dt}&=-\gamma x + I(t)x(1-x)
\end{align}
where $\gamma$ is the damping parameter, $x$ is the node state and $I(t)$ is the time-dependent drive.

\begin{figure}[h]
\centering
\includegraphics[width=0.9\textwidth]{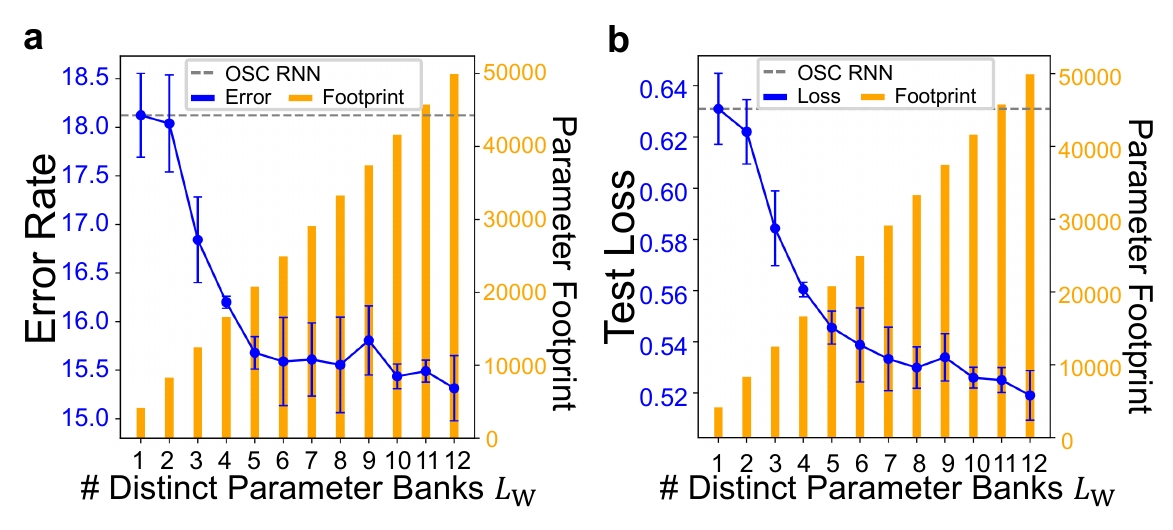}
\caption{\textbf{|TIDAL-Net performance with Oscillator Network performance.} $\mathbf{a}$: Error Rate $\mathbf{b}$: Test Loss (Crossentropy) Like TIDAL-Net with ReLU, the error and loss (blue plot) decrease as the number of parameters (orange bars) increase. This means that TIDAL-Net with Oscillator Network has the same advantages that TIDAL-Net with ReLU has.}\label{fig:OSC}
\end{figure}

Supplementary Figure~\ref{fig:OSC} shows the results of the numerical experiment of TIDAL-Net with this Oscillator Network on SVHN classification tasks.
The number of training epochs was $900$.
Generally speaking, we observe the same trends as the results for a more traditional ReLU network shown in the main manuscript Fig.~\ref{fig:svhn_results}.
This shows that TIDAL-Net is not limited to a particular activation function.

\clearpage

\section{MNIST classification}
As explained in the main text, we used SVHN as a dataset instead of MNIST, which is arguably the most common benchmark for image classification in PNNs.
MNIST is a relatively simple benchmark, for which a large fraction of the classification performance can already be achieved by simple or nearly linear models.
We therefore found it insufficient for consistently showing trends and performance differences, as many models saturate performance.
For reference, however, we show the results with MNIST in this section.

\begin{figure}[h]
\centering
\includegraphics[width=0.9\textwidth]{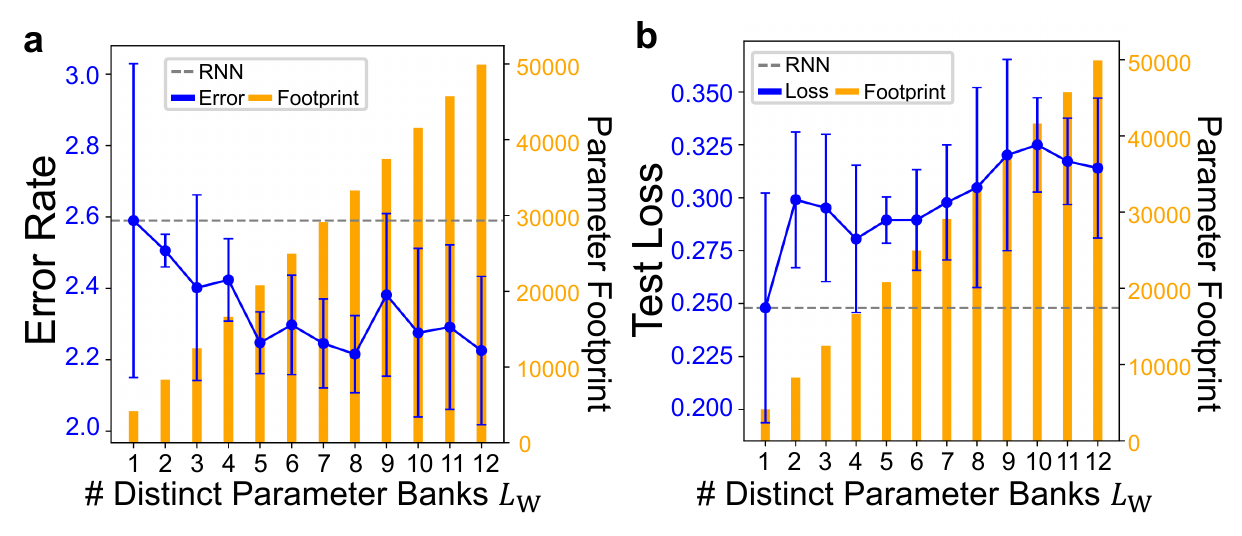}
\caption{\textbf{|TIDAL-Net performance on image classification tasks with MNIST.} 
The number of weights $L_{\mathrm{W}}$ is changed while keeping the number of layers $L_{\mathrm{T}}$ constant. 
$\mathbf{a},\mathbf{b}$ Error Rate and Parameter Footprint.
Blue curves show the mean performance, error bars indicate the standard deviation across trials, and orange bars show the number of trainable hidden-layer parameters, corresponding to the hardware footprint.
The gray dashed line is the performance of the stateless RNN limit for $L_\mathrm{W} = 1$.
While TIDAL-Net can outperform standard RNNs with additional layers like Fig. \ref{fig:svhn_results} on the errors, the performance increase is small.
Especially, the error bars are too large to distinguish the performance.
}\label{fig:mnist_results}
\end{figure}

Supplementary Figure \ref{fig:mnist_results} shows the results of the MNIST image classification. 
Similar to the results in Fig.~\ref{fig:svhn_results}, TIDAL-Net outperforms standard RNNs with additional layers in Fig.~\ref{fig:mnist_results}{$\mathbf{a}$}.
However, the value improvement in the error rate is small.
Furthermore, the error bars for both the error rate and the test loss in Supplementary Fig.~\ref{fig:mnist_results} are large relative to the differences among the models.
Since the standard RNN has already achieved quite low error and loss values, we consider that MNIST is too easy to highlight the performance differences in our settings.
Furthermore, reducing the error rate from $\leq3\%$ to $\leq 1\%$ is often only a question of finetuning data augmentation methods.

\clearpage

\section{Implementation details of TIDAL-Net}\label{section_method} 
In this section, we describe the details of the Time-Indexed Deep Alternating Layers Network (TIDAL-Net) framework and explain its practical advantages in implementing PNNs. 
The following subsections detail each component of our method.

\subsection{Intuition}\label{subsection_intuition}
Switching network weights across time steps can increase the number of parameters. 
For example, doubling the number of distinct weight matrices used in the recurrent loop effectively doubles the number of parameters, even without increasing the width of the layers. 
Simultaneously, switching between separate weight banks - each encoding a unique connectivity between nodes - introduces a more complex computational structure. 
Stacking networks that have different connections is widely known to improve representation power even with limited parameters \cite{montufar2014number}.
Furthermore, from a hardware perspective, switching network weights may be significantly easier than duplicating the full network structures, particularly in the context of PNNs, where component size and reconfigurability are the key limitations. 

The TIDAL-Net thus represents a middle ground between a stateless RNN and a fully time-variable DNN, where its weights change over time as Supplementary Table \ref{tab_RNN} shows.
As such, TIDAL-Net offers a practical and scalable solution for enhancing neural network expressivity while maintaining feasibility in physical implementation.

\begin{table}[h]
\caption{\textbf{|Comparison among RNNs.} TIDAL-Net exists between a stateless RNN and a time-variable DNN.}\label{tab_RNN}%
\begin{tabular}{@{}llll@{}}
\toprule
 & stateless RNN & TIDAL-Net  & time-variable DNN\\
\midrule
input/output weight    & fixed   & fixed  & fixed  \\
network weight    & fixed   & periodic switching  & dynamical switching \\
\botrule
\end{tabular}
\end{table}

\subsection{Back-Propagation Through Time (BPTT)}\label{subsection_BPTT}

In supervised machine learning, the training dataset consists of pairs of examples and targets $\{(x^{k}, y_{\rm{target}}^{k})\}_{k=1}^{N_{\rm{batch}}}$, where $N_{\rm{batch}}$ is the batch size.
During training, the goal is to obtain parameters of the network that maps any given example $x^{k}$ to an output (prediction) $\hat{y}^{k}$ that is as close as possible to the desired target $y_{\rm{target}}^{k}$.
In gradient-based optimization, this objective is achieved by minimizing a loss function $L_{\rm{batch}}$ over the dataset.
\begin{align}
L_{\rm{batch}} &= \frac{1}{N_{\rm{batch}}}\sum_{k} L^{k}
\end{align}
where $L^{k}=l(y_{\rm{target}}^k,\hat{y}^k)$ is the loss on the $k$-th data and $l$ denotes the chosen loss function. 
In our experiments, we use the cross-entropy loss $l$, which is explained in detail in subsection \ref{subsection_cross_entropy}.

Once the loss gradient with respect to model parameters is computed, parameters can be updated using any gradient descent-based optimizer. 
In our study, we use the Adam optimizer \cite{adam}, which is introduced in subsection \ref{subsection_Adam}, with a learning rate of $0.001$.

To train TIDAL-Net, we apply Backpropagation Through Time (BPTT), which accounts for dependencies across time steps.
BPTT computes the gradient of the loss with respect to the parameters $\phi=(\alpha,W_{\rm{xh}},W_{\rm{hh}}^1,...,W_{\rm{hh}}^p,...,W_{\rm{hh}}^P,b_{\rm{h}}^1,...,b_{\rm{h}}^p,...,b_{\rm{h}}^P,W_{\rm{hy}},b_{\rm{y}})$ as:
\begin{align}
\frac{\partial L^k}{\partial \phi}&=\sum_{t=1}^T\frac{\partial L^k}{\partial h[t]}\frac{\partial h[t]}{\partial \phi}\\
&=\sum^T_{t=1}\sum_{i=1}^t\frac{\partial L^k}{\partial h[t]}\cdot\left(\prod_{j=i+1}^{t}\frac{\partial h[j]}{\partial h[j-1]}\right)\cdot\frac{\partial h[i]}{\partial \phi}
\end{align}
This recursive computation captures the temporal dependency of the hidden states across multiple time steps.

\subsubsection{Chain rule through time}
Based on the update equation~ (\ref{eq:relax_net}), each $h[t]$ is computed recursively from its previous state $h[t-1]$.
In our implementation using PyTorch, the computational graph is automatically constructed during the forward pass. This graph stores: (1) intermediate values (like $h[t]$), (2) dependencies (how $h[t]$ is computed from $h[t-1]$, weights, and inputs).
As a result, PyTorch can automatically compute partial derivatives such as $\frac{h[j]}{h[j-1]}$ using autograd.

Using the chain rule through time, the gradient of the loss with respect to each intermediate hidden state can be expressed as:
\begin{align}
\frac{\partial L^k}{\partial h[t]}
&=\frac{\partial L^k}{\partial h[T]}\cdot\left(\prod_{j=t+1}^{T}\frac{\partial h[j]}{\partial h[j-1]}\right)
\end{align}
where $\frac{\partial L^k}{\partial h[T]}$ can be directly computed from the loss function in our experiments.
So, it is possible to get $\frac{\partial L^k}{\partial h[t]}$.
In addition, $\frac{\partial h[t]}{\partial\phi}$ can be computed by autograd.
Thus, BPTT effectively propagates the gradients backward through time, allowing parameter updates via gradient-based optimization.

Nevertheless, TIDAL-Net is also compatible with alternative training methods such as DFA-based \cite{nakajima2022physical} or adjoint-based methods \cite{sunada2025}.

\subsection{Cross entropy}\label{subsection_cross_entropy}
Cross-entropy is a standard loss function used in classification tasks, which measures the dissimilarity between a true distribution $\mathbf{p}$ and an estimated distribution $\mathbf{q}$.
In most classification settings, the ground truth $\mathbf{p}$ is represented as a one-hot encoded vector, and $\mathbf{q}$ is the probability distribution output by the model (e.g., through a softmax layer). 
For discrete distributions $\mathbf{p}$ and $\mathbf{q}$ over $C\in\mathbb{N}$ classes, where $p_c$ and $q_c$ express the $\mathbf{p}$ and $\mathbf{q}$ probability on the class $c=1,...,C$, the cross-entropy is computed as:
\begin{align}
H(\mathbf{p},\mathbf{q})&=-\sum_{c=1}^C p_c \log{q_c}
\end{align}

For the image classification task (SVHN), we set $C=10$ (digits $0$ through $9$). (see subsection \ref{subsection_svhn_classification} for SVHN dataset)
For the next-token prediction task (Shakespeare corpus), we use $C=43$, which corresponds to the number of unique tokens after preprocessing. (see subsection \ref{subsection_next_token_prediction} for Shakespeare corpus dataset)
This loss encourages the model to assign high probability to the correct class while penalizing incorrect predictions.

\subsection{Adaptive moment estimation (Adam)}\label{subsection_Adam}
Adam \cite{adam} is a common optimizer for deep learning.
It combines the benefits of two other popular methods: momentum \cite{polyak1964momentum} and RMSProp \cite{hinton2012rmsprop}.
Given a parameter $\psi_t$, loss gradient $g_t=\nabla_{\psi} L_\mathrm{T}$ at time step $t$,
Adam uses two kinds of moments:\\
First moment (mean):
\begin{align}
m_t &= \beta_1 m_{t-1}+(1-\beta_1)g_t
\end{align}
Second moment (uncentered variance):\\
\begin{align}
v_t &= \beta_2 v_{t-1} +(1-\beta_2)g_t^2
\end{align}
where $\beta_1, \beta_2$ are exponential decay rates for the moving averages.
The first moment $m_t$ smooths the gradient similarly to momentum, while the second moment $v_t$ adapts the learning rate similarly to Root Mean Square Propagation (RMSProp).

To correct the bias introduced during initialization, Adam performs the following bias corrections:
\begin{align}
\hat{m}_t&=\frac{m_t}{1-\beta_1^t}\\
\hat{v}_t&=\frac{v_t}{1-\beta_2^t}
\end{align}

Finally, the parameters are updated as:
\begin{align}
\theta_{t+1}&=\theta-\alpha_{\rm{adam}}\cdot\frac{\hat{m}_t}{\sqrt{\hat{v}_t}+\epsilon} 
\end{align}

In our experiments, we used $\alpha_{\rm{adam}}=0.001,\beta_1=0.9,\beta_2=0.999,\epsilon=10^{-8}$, which are default values for Adam. 

\subsection{Initialization of the hidden layer parameters}\label{subsection_initialization}
Deep neural networks are often susceptible to the problems of vanishing and exploding gradients, which can severely affect convergence during training.
Appropriate initialization of weights plays a crucial role in maintaining the stability of the forward and backward signal propagation across layers.

In this study, we employ a widely used initialization strategy depending on the choice of activation function for a layer with $\lambda_{\rm{in}}$ inputs and $\lambda_{\rm{out}}$ outputs:

Kaiming Initialization (He Initialization) \cite{he2015delving}:
\begin{align}\label{eq_kaiming}
w_{ij}\sim U\left(-\sqrt{\frac{6}{\lambda_{\rm{in}}}},\sqrt{\frac{6}{\lambda_{\rm{in}}}}\right)
\end{align}
where $w_{ij}$ expresses the values of weight matrices.
This scaling preserves the variance of activations through ReLU layers, and is appropriate for models with ReLU as an activation function.
In this study, this method is used because ReLU is selected as the activation function.

The strength of the residual connection, which is not a weight parameter, is an exception at the point of initialization. 
It is set to $1.0$ at first, and it is changed during training.

\subsection{Validation dataset}\label{subsection_validation}
To evaluate and tune our model, we created a validation dataset by splitting the original training dataset into two parts using a $9:1$ ratio. 
The larger portion was used to train the model, while the smaller portion was reserved for validation purposes. 
This validation dataset is used in tuning hyperparameters, such as the number of hidden units in a layer or the number of weight switching steps in the TIDAL-Net structure.
The resulting training dataset was also shuffled before use in mini-batch training, which is described in subsection \ref{subsection_mini_batch}.

\subsection{Mini-batch gradient descent}\label{subsection_mini_batch}
As we explained in subsection \ref{subsection_Adam}, we train the model with BPTT, which is a kind of a gradient-based (gradient descent) method.
At the same time, we use Mini-batch Gradient Descent, which is the middle ground between Gradient Descent and Stochastic Gradient Descent (SGD). 
Gradient Descent computes the exact gradient of the loss over the entire training set, then takes one parameter update. 
While precise, this approach can be computationally intensive, especially for large datasets. 
SGD instead approximates that gradient by computing it on a single random example and immediately updating the model. 
Although this results in faster updates, it introduces significant variance and noise into the optimization process.
Mini-Batch Gradient Descent serves as a compromise between these two methods. 
In our study, we use mini-batches of size $N_{\rm{batch}}$ for both training and evaluation on the SVHN classification task. 
Each update step is based on the gradient computed over one mini-batch. 
This method offers the computational efficiency of SGD while reducing the variance in gradient estimates, which contributes to more stable and effective learning. 
Additionally, mini-batch processing is well-suited to modern parallel hardware architectures, making it the default choice for training deep models.

\subsection{SVHN classification}\label{subsection_svhn_classification}
In this subsection, we describe the experimental setup used to evaluate TIDAL-Net on an image classification task.
The objective of this experiment is to demonstrate how TIDAL-Net can improve the performance of a recurrent neural network compared to a conventional RNN.

We train the model using the Street View House Numbers (SVHN) dataset \cite{svhn}, which contains over $600000$ color images of digits ($0$–$9$) extracted from real-world Google Street View photographs.
Specifically, the dataset includes $73257$ images for training and $26032$ images for testing. 
Each image is a cropped digit, making this a ten-class classification problem.

Among image classification tasks, MNIST (Modified National Institute of Standards and Technology) \cite{lecun1998gradient} may be the most popular dataset because of its simplicity.
Compared to MNIST, the SVHN dataset poses a more challenging classification task due to its real-world complexity, color variations, and background clutter. 
We selected SVHN over MNIST to better assess the expressivity and performance improvements provided by the TIDAL-Net framework under more practical conditions. 

To avoid over-fitting, we used data augmentation techniques on the SVHN dataset. 
These techniques are explained in subsubsection \ref{subsubsection_randomcrop}, \ref{subsubsection_color_jitter}, \ref{subsubsection_gaussian_noise} and \ref{subsubsection_normalize}.
Then, we explain how we train TIDAL-Net with SVHN in \ref{subsubsection_svhn_training}

\subsubsection{Randomcrop and padding}\label{subsubsection_randomcrop}

Each image in the SVHN dataset is a $32\times32\times3$ RGB image stored as unsigned 8-bit integers. 
To introduce spatial variation and improve generalization, we first pad each image by $4$ pixels on all sides, expanding its dimensions to $40\times40$.
After padding with pure black that all components are $0$, we extract a randomly located $32\times32$ crop from the padded image. 
This augmentation simulates minor translations as well as zoom-in and zoom-out effects, encouraging the model to learn spatial invariance in its representations.

\subsubsection{Color jitter}\label{subsubsection_color_jitter}
Following spatial augmentation (random crop and padding), we apply random color jittering to simulate changes in lighting and scene conditions. Specifically, we perturb the brightness, contrast, and saturation of each image by up to $\pm10\%$, and adjust the hue by up to $\pm5\%$.
These transformations help the model become robust to variations commonly observed in real-world street imagery. 
Given that the SVHN dataset is derived from natural photographic scenes, such augmentations serve to better align the training distribution with the expected variability.

\subsubsection{Add gaussian noise}\label{subsubsection_gaussian_noise}
We convert SVHN values from $[0, 255]$ integer to $[0, 1]$ float values by dividing them by $255$ with converting it to a tensor.
To further improve generalization, we inject pixel-wise Gaussian noise (standard deviation of the noise is $0.005$).
This augmentation encourages the model to become invariant to minor artifacts such as compression noise or small, irrelevant details and helps to reduce the risk of overfitting.

\subsubsection{Normalize}\label{subsubsection_normalize}
After applying noise, we normalize each RGB channel per image to have zero mean and unit variance, assuming the inputs are already scaled to the $[0,1]$ range. 
This normalization procedure accelerates convergence by ensuring consistent gradient flow across different input dimensions. 
Finally, each image, originally a $32\times32\times3$ tensor, is reshaped into a $3072$-dimensional vector. 
These vectors are then used as inputs to the TIDAL-Net, repeatedly.

\subsubsection{SVHN training}\label{subsubsection_svhn_training}
Then, we train the models with BPTT stated in subsection \ref{subsection_BPTT}.
As we explained in subsection \ref{subsection_BPTT}, the input is expressed as a vector $x^k[t]\in\mathbb{R}^X$ where $k$ is the image (data) index in the mini-batch data, whose size is $N_{\rm{batch}}$. 
Here, $N_{\mathrm{batch}}$ is the number of examples processed together in one optimizer step.
In this image classification task, we input the same image $x_{\rm{image}}$ repeatedly. (In other words, $x^k[t]=x_{\rm{image}}=\rm{const}$ for $t=0,...,L_{\rm{T}}-1$.)

As we use the output $\hat{y}^k$ from the equation~(\ref{eq:time-multiplicated_output}) as the prediction result in the numerical experiments,
\begin{align}
L_k&=l(y^k_{\rm{target}},\hat{y}^k)
\end{align}

Then, we trained the model with Adam. 

\subsection{Next token prediction}\label{subsection_next_token_prediction}
One of the main reasons RNNs are widely studied is their natural ability to process sequential data \cite{graves2013generating}. 
In this section, we evaluate TIDAL-Net on a time-series prediction task, specifically next-token prediction. 
Although a deep feedforward network may work with sequential data by simultaneously inputting all tokens as part of a large input vector, TIDAL-Net is structurally distinct due to its recurrent dynamics and weight-switching mechanism. 
We therefore employ a different input strategy that befits TIDAL-Net, which differs from just repeating the DNN as explained in section \ref{section_repeated_DNN}.
This difference makes it particularly interesting to study how TIDAL-Net behaves in sequential tasks compared to conventional models.

Next-token prediction involves learning a model that, given a sequence of past inputs, predicts the most likely next symbol or token in the sequence. 
This task captures the temporal dependencies in the data and is commonly used in language modeling and other autoregressive applications.

\subsubsection{Shakespeare dataset}\label{subsubsection_shakespeare}
For the next-token prediction task, we use a dataset derived from the Folger Shakespeare Library’s digital editions \cite{folger_shakespeare}. 
The dataset consists of plain-text versions of Shakespeare's plays, containing a total of $9044939$ characters. 
To convert the text into a trainable format suitable for RNN input, we applied a series of preprocessing and tokenization steps, described in subsubsection \ref{subsubsection_tokenization}.

\subsubsection{Tokenization}\label{subsubsection_tokenization}
Before tokenization, the text is first preprocessed to simplify further processing. 
All uppercase letters are converted to lowercase, and variant characters are mapped to their modern equivalents in the English alphabet. 
Infrequent characters, defined as those with an occurrence rate below $0.01\%$, are replaced with a special placeholder symbol to limit the vocabulary size.

Next, we construct a vocabulary by counting the frequency of all fixed-length substrings (n-grams) within the full corpus. 
In this experiment, we set $n=1$, so the tokens correspond to individual characters. 
Then, we can build vocabulary by giving an index to the tokens.
The placeholder character is assigned index $0$ in the vocabulary. 
All other characters are assigned indices based on frequency, with the more frequent characters given lower index values.

The final vocabulary contains all characters that exceed the minimum frequency threshold. 
Using this vocabulary, we tokenize the Shakespeare text.
For each position in the text, we identify the character that exists in the vocabulary. 
If a match is found, the corresponding token is emitted, and the pointer is advanced. 
If no valid match is found, the placeholder token is emitted, and the pointer advances by one character. 
Finally, we slice the integer array into a contiguous $90\%$ train / $10\%$ validation split.

That integer array is reshaped into fixed‐length windows of size $T+1$.
The length of each sample is $T+1$. 
Inputs are the first $T$ tokens, and targets are the last $T+1$-th token.
These (input, target) pairs are what the TIDAL-Net actually consumes during training.

\subsubsection{Time series input}\label{subsubsection_time_series}
The input can be expressed as a sequence $\mathbf{x}^k=(x^k[1],..x^k[T]),\ x^k[t]\in\mathbb{R}^X$ where $k$ is the data index in the mini-batch data. 
In this next token prediction task, we embed the token scalers $x[t]$ to embedding vectors $x_{\rm{e}}[t]\in\mathbb{R}^{36}$ before inputting them into RNN as an input $\mathbf{x}_{\rm{e}}^k=(x^k_{\rm{e}}[1],..x^k_{\rm{e}}[T])$.

As in the image classification task, we obtain the output $\hat{y}^k$ from the equation~(\ref{eq:time-multiplicated_output}) as the prediction result in the numerical experiments
\begin{align}
L_k&=l(y^k_{\rm{target}},\hat{y}^k)
\end{align}

Then, we trained the model with Adam. 

\subsubsection{Variational RNN}\label{subsubsection_varidational_RNN}

Unlike image-based tasks, textual data does not lend itself to spatial augmentations. 
To mitigate overfitting in the next-token prediction task, we apply a regularization technique known as variational dropout \cite{gal2016theoretically}, which is specifically adapted for RNNs.

Unlike standard dropout, where dropout masks are independently resampled at every time step, variational dropout applies fixed dropout masks across both the temporal and depth (layer) dimensions. 
This preserves temporal consistency and prevents information leakage that may arise from fluctuating dropout patterns during sequence modeling.

To implement variational dropout, on each mini-batch dataset, we generate three kinds of dropout masks per mini-batch: one each for the input-to-hidden $M_{\rm{xh}}^{\rm{dp}}$, hidden-to-hidden $M_{\rm{hh}}^{\rm{dp}}$, and hidden-to-output $M_{\rm{hy}}^{\rm{dp}}$ connections.
These masks are sampled from a Bernoulli distribution and applied as multiplicative noise before the corresponding matrix operations as follows:
\begin{align}
x^{\rm{dp}}[t]&:=M^{\rm{dp}}_{\rm{xh}}x[t]\\
h^{\rm{dp}}[t]&:=M^{\rm{dp}}_{\rm{hh}}h[t]\\
\hat{y}^{\rm{dp}}&:=W_{\rm{hy}}(M^{\rm{dp}}_{\rm{hy}}h[L_{\rm{T}}]+ b_{\rm{y}})
\end{align}
Using them instead of $x[t],h[t],\hat{y}$, the update equation~(\ref{eq:time-mutliplexed-DNN}) can be rewritten: 
\begin{align}
   h^{\rm{dp}}[t] &= M_{\rm{hh}}^{\rm{dp}}(\alpha h^{\rm{dp}}[t-1] + f (W_{\rm{xh}}x^{\rm{dp}}[t]+ W_{\rm{hh}}[t]h^{\rm{dp}}[t-1] + b_{\rm{h}}[t]))
\end{align}
The consistent masking across time steps ensures stable dynamics while promoting generalization.

Although this dropout method in variational RNN makes it possible to prevent overfitting, it also results in performance increase saturations at the same time.

\subsubsection{NLP performance decrease with large number of distinct weights $L_{\rm{W}}$}
In Fig.~\ref{fig6:NLP}, TIDAL-Net shows a different trend on results on a NLP task compared to image classification.
Strikingly, the error rate first decreases with the number of distinct weights $L_{\rm{W}}$, as expected, but then when $L_{\rm{W}}$ is large it increases again.
This is the opposite of the expected effect, as more parameters generally imply better performance for neural networks.
To further investigate the details of next token prediction tasks, we show the change of loss changing during the training epochs.

\begin{figure}[h]
\centering
\includegraphics[width=0.6\textwidth]{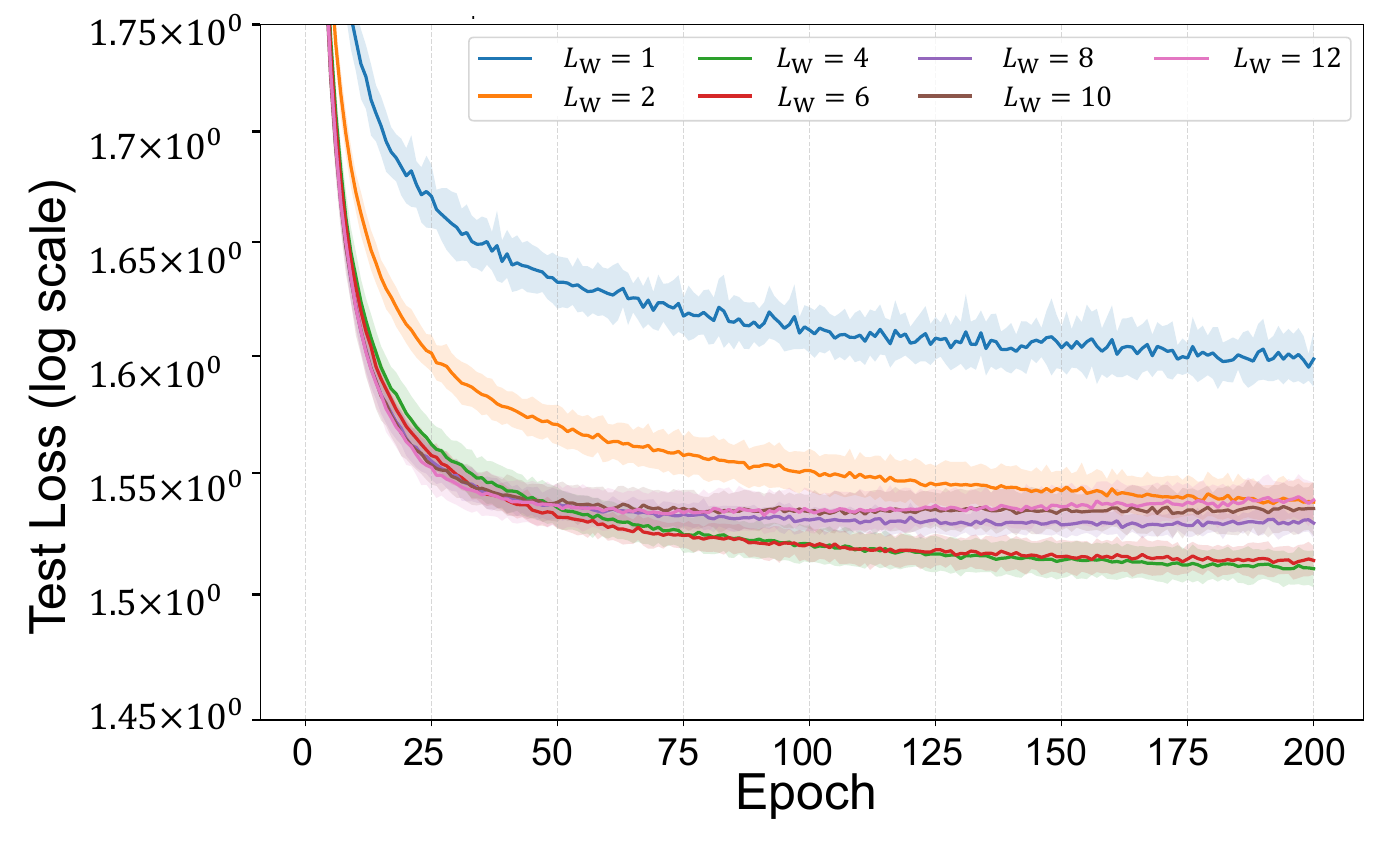}
\caption{\textbf{| Epochs vs TIDAL-Net performance on NLP.} Test Loss changes as epoch increases. The loss values generally decrease as training progresses. The loss learning curves flatten out at higher values as $L_{\rm{W}}$ increases, which is likely caused by the effect of gradient vanishing/exploding. }\label{fig:nlp_epoch}
\end{figure}

Supplementary Figure~\ref{fig:nlp_epoch} shows the change of test loss on the different $L_{\rm{W}}$.
We prepare $5$ trials, and the pastel colors show the standard deviations.
The epoch $200$ values are the results that Fig.~\ref{fig6:NLP} show.
As can be seen, the test loss flattens at a higher value when $L_{\rm{W}}$ increases (compare pink and red lines in Supplementary Figure~\ref{fig:nlp_epoch}).
The flat shape indicates that TIDAL-Net cannot increase its performance with large epochs.
Generally speaking for such characteristics, there are some possible causes: insufficient dataset, inappropriate parameters (learning rate, initialization), dropout effect and gradient vanishing/exploding etc.
Considering that the performance on the $L_{W}=6$ case is better than the ones on larger $L_{W}$ cases, there should be enough data, appropriate parameters, and a dropout setting.
Thus, gradient vanishing/exploding is considered to be the cause of the performance decrease.

In particular, with $L_{W}$ sufficiently smaller than the number of temporal steps (here: $L_\mathrm{T} = 12$), every layer is repeated closer to the output section, allowing it to be trained even if gradients vanish. 

\subsection{Repeating DNNs and TIDAL-Net}\label{section_repeated_DNN}
TIDAL-Net shares structural similarities with models that are constructed by repeating deep neural networks.
In principle, repeatedly applying a small DNN block can produce a sequence of transformations similar to that of TIDAL-Net.
If the number of distinct weights $L_\mathrm{W}$ divides the number of time steps $L_\mathrm{T}$, the TIDAL-Net update can be viewed as repeated applications of an $L_\mathrm{W}$-step block.
In other words, the model resembles a stateless RNN whose recurrent unit is an $L_\mathrm{W}$-step block.
At this coarse-grained level, the block itself plays the role of a recurrent transformation.

However, TIDAL-Net is constructed to also handle cases where layers are only partially repeated. 
Similarly, we allow the input to be injected at different time steps. 
This allows TIDAL-Net to handle sequential inputs over time, rather than requiring all input data to be presented simultaneously. 
This makes TIDAL-Net suitable for time-series tasks, where inputs arrive incrementally, such as the NLP task shown in the main manuscript.

Note, in the field of photonic computing, repeating simple layers has been proposed and physically demonstrated, which are based on Diffractive Deep Neural Networks (D${^2}$NN) \cite{lin2018all, zhou2024recurrent}. 
For TIDAL-Net, we would consider the sets of diffractive layers of a D${^2}$NN as a single hidden layer, as the nonlinear function is only applied once at the read-out.
In that sense, D${^2}$NN can be considered to have demonstrated the case of a stateless RNN with $1=L_\textrm{W}<L_\textrm{T}$.
These platforms are promising candidates for implementing a full TIDAL-Net like structure with several distinct layers.

In general, in TIDAL-Net, only the network weights need to be switched or cycled during operation, so that implementations of the activation function can be reused as much as possible (which also increases the uniformity of the nonlinear activation function across layers). 

Consequently, TIDAL-Net can be viewed as an extended and more physically feasible variant of repeated DNN architectures. 
In this paper, we focused on the simplest versions of a TIDAL-Net structure, with a periodic repetition structure. 
In principle, there is no need to choose this particular order, and aperiodic structures could also be used, which would not be possible by merely repeating a small multi-layer DNN several times.

\subsection{Training in real experiments}
TIDAL-Net can be trained using existing training strategies for PNNs. 
In the numerical experiments in the main manuscript, we unroll the temporal dynamics and train the parameters using standard backpropagation through time as explained in the previous subsections. 
The trainable parameters are the input and output weights, the residual coefficient, and the finite set of preconfigured hidden-layer weight banks and biases. 
In an experimental implementation, one possible route is to train a calibrated differentiable model of the physical system in silico and then program the obtained parameters into the physical device. Another route is hardware-in-the-loop training, such as physics-aware training, in which the forward response is measured from the physical system while parameter updates are computed using a differentiable surrogate model or experimentally estimated gradients. 
Importantly, TIDAL-Net does not require the trainable physical weights to be reconfigured at the inference timescale. 
The weights can be updated on the slower training or calibration timescale, while inference only requires fast switching among the trained weight banks.

\clearpage

\section{Reference performance levels for SVHN image classification}
In the main text, we present TIDAL-Net’s performance on the SVHN image-classification task~\cite{svhn}. 
For reference, we provide representative performance levels from related models and baselines reported on the same dataset.

In Supplementary Table~\ref{tab_svhn_comparison}, we present performance examples for the SVHN image classification task.
In the case of Fit-DNN \cite{stelzer2021deep} ($N=100$, $L=2$, which are parameters defined in \cite{stelzer2021deep}), the number of trainable parameters and the achieved accuracy of the Folded-in-time Deep Neural Network (Fit-DNN) are comparable to those of the Standard RNN and TIDAL-Net.
In the original study proposing SVHN \cite{svhn}, Linear Support Vector Machine (SVM) with Histogram of Oriented Gradients (HOG), which generates characteristic features, also demonstrated a similar performance.
These results indicate that TIDAL-Net achieves a reasonable performance level compared with both a PNN (Fit-DNN \cite{stelzer2021deep}) and a conventional reference model used in the SVHN benchmark (Linear SVM with HOG \cite{svhn}).

\begin{table}[h]
\caption{\textbf{|SVHN classification performance comparison.} TIDAL-Net achieves reasonable performance among other previous approaches. Models have different architectures separated into different components. ``Input (Hidden, Output)" expresses the number of parameters in the input weight (hidden layers, output weight) as a multilayer perceptron. ``Other" expresses the number of parameters coming from other components, such as the strength of residual connections. In addition, there are blanks in Human because we do not know how many parameters are used in the Human brain for image classification.}\label{tab_svhn_comparison}%
\begin{tabular}{@{}l|rrrrr@{}}
\toprule
Model   & SVHN Accuracy & Input & Hidden & Output & Other \\
\midrule
Fit-DNN ($N=100$, $L=2$) \cite{stelzer2021deep} & 78.9\% & 102,500 & 10,100 & 1,010 & - \\
Standard RNN  & 81.2\% & 196,672 & 4,160 & 650 & 1\\
TIDAL-Net ($L_{\rm{W}}=12$)  & 86.1\%  &  196,672 & 49,920 & 650 & 1\\
Linear SVM with HOG \cite{svhn} & 85.0\% & - & - & - & 1765 \\
Human \cite{svhn} & 98.0\% &  &  &  &  \\
\botrule
\end{tabular}
\end{table}

\clearpage

\section{Examples of next token prediction}
We evaluated TIDAL-Net on a next-token prediction task.
A trained TIDAL-Net can be used as a generative model with input prompts.
To illustrate the qualitative behavior of the trained model, we show generated character sequences at different training stages.

Supplementary Table \ref{tab_NLP} shows $20$-character (here, a space is seen as one character) output sequences generated with the input prompt “to be, or not to be,” at different training epochs under the same setting as Fig.~\ref{fig6:NLP}.
The trained TIDAL-Net model generates successive characters following the given input. 
As training progresses, the outputs evolve from incomprehensible strings to increasingly reasonable sequences. 
This trend is the same among different $L_{\rm{W}}$ cases.

\begin{table}[h]
\caption{\textbf{|Output examples on NLP tasks.} TIDAL-Net outputs when using TIDAL-Net as a generative model with the input prompt ``to be, or not to be," from the Fig.~\ref{fig6:NLP} setting. As training progresses, more coherent continuations appear. Note, that our models are far smaller than even GPT-2 and thus cannot be expected to produce full sentences consistently.}\label{tab_NLP}%
\begin{tabular}{@{}l||llll@{}}
\toprule
 epoch & $L_{\rm{W}}=1$ & $L_{\rm{W}}=4$ & $L_{\rm{W}}=8$ & $L_{\rm{W}}=12$\\
\midrule
$1$ & aiuer rol ruitam r- & me sous the the the & tttttttttttttttttttt & are                 
\\
$100$ & and thoopple they so
 & and the state the st
 & akiccxeanxtny top th
 & and the state the st
\\
$200$ & and thee the king pe
 & and the stranger the
 & akicletn tay tpofray
 & my lord, and the cou
\\
\botrule
\end{tabular}
\end{table}

\clearpage

\section{Parameter counts of existing digital and physical neural networks}

In the Introduction of the main manuscript, we discussed the importance of the number of trainable parameters for machine learning.
Here, we provide the sources that were used in the creation of Fig.~\ref{fig:FIGINTRO}.

First, we discuss how we obtained the parameter information of models for digital electronic computers.
All parameter counts are based on values reported by the original authors or developers of the respective models. 
We did not include models for which we could not identify an official or clearly documented parameter count.
Note that this excludes some of the most advanced Large Language Models (LLMs), which are most likely the largest systems, meaning that the true gap between Physical Neural Networks (PNNs) and digital neural networks might be even larger than shown.
For AlexNet, the authors report that their model has $60$ million parameters \cite{krizhevsky2012imagenet}.
For the VGG19 paper \cite{simonyan2014very}, their model was reported to have $144$ million parameters.
The GPT-1 paper reports that the GPT-1 model has $117$ million trainable parameters \cite{radford2018improving}, including $31$ million from embedding, $85$ million from $12$ Transformer blocks \cite{Transformer}, and $0.5$ million in LayerNorms and positional embedding.
The authors of BERT-Large report that their model has $340$ million parameters \cite{devlin2019bert}.
Megatron-LM has been described in the paper to have $8.3$ billion parameters \cite{shoeybi2019megatron}.
For GPT-2, OpenAI researchers reported that they have achieved a $1.5$ billion parameter model \cite{radford2019language}.
The T5-11B researchers trained a $11$ billion parameter model according to the main paper \cite{raffel2020exploring}.
GPT-3 is reported as having $175$ billion trainable parameters \cite{brown2020language}.
In the development of the Switch Transformer, researchers designed it to have $1.6$ trillion parameters \cite{fedus2022switch}.
The PaLM model was trained with $540$ billion parameters \cite{chowdhery2023palm}.

The estimation of parameters in PNN models can sometimes be more difficult.
We made a careful effort, but not all systems are straightforwardly interpretable. 
Here, we explain the source of the reported number of parameters of each PNN model.
Note that all PNN systems that we cite require electronic supporting components. 
In this sense, they are hybrid systems. 
Supplementary Table \ref{tab_previous_research} lists the PNNs and their sizes, where we have categorized the PNNs based on their primary physical mechanism.
Note that, crucially, we do not distinguish between different levels of noise or resolution, e.g., systems using 4-bit vs 8-bit or 10-bit resolution in their trainable parameters are counted the same way. 
We do not believe that this is necessarily always justified, but in the absence of a coherent theory of physical computing \cite{jaeger2023toward}, we will treat all these parameters as equivalent for now.

In Reservoir Computing (RC), the only trainable part is the output weight matrix. 
This means that the number of trainable parameters depends on the output dimension of the reservoir.
Assuming a scalar variable per reservoir node, the number of parameters equals the number of reservoir nodes.
Crucially, it is known that the computing capability of a RC is decided by the number of nodes \cite{dambre2012information}.
The number of parameters of the RC systems in Supplementary Table \ref{tab_previous_research} were taken from the following references:
A Photonic RC with $2025$ nodes \cite{bueno2018reinforcement}, an optoelectric RC with $400$ virtual nodes \cite{appeltant2011information}, an integrated photonics RC with $16$ nodes \cite{vandoorne2014experimental}, a Spin-Torque Oscillator RC with $400$ nodes \cite{torrejon2017neuromorphic}, a MEMS Nonlinear Resonator RC with $440$ nodes \cite{sun2021novel} and an electrochemical RC with $112$ nodes \cite{kan2022physical}.

For the Diffractive Deep Neural Network (D${}^2$NN), a neuron corresponds to a point of a diffractive layer with a phase-only transmission coefficient, which is learnable.
Accordingly, the number of trainable parameters corresponds to this number of neurons.
The paper cited implements $0.45$ million neurons in the system \cite{lin2018all}, one of the largest size PNN demonstrations to this day.
Physical Aware Training (PAT) \cite{wright2022deep} is not limited to photonics, but the largest system in the original paper used photonics and achieved $50568$ parameters.
The shown Nanophotonic Neural Network had $56$ Mach-Zehnder Interferometers \cite{shen2017deep}, where each of them had $2$ parameters (an internal and external phase) for a total of $112$ trainable parameters.
For the Micro-Ring Resonator (MRR) RNNs, a total of $576$ MRR weights are used \cite{tait2017neuromorphic}.
``Mars" by Lightmatter can achieve $64\times64$ matrix multiplications, implying that the system has $4096$ parameters \cite{ramey2020silicon}.
The parameters of the DOPO Network are implemented in a separate FPGA, and the number of learnable values is $57600$ \cite{inagaki2021collective}.
For the Coupled Spin Oscillator Network, $4$ oscillators are achieved, with each of them having a bias current, which is the trainable part.
In addition, there are $26$ parameters in input weights as software components, for a total of $30$ parameters \cite{romera2018vowel}.
In the Magnetoresistive Random-Access Memory (MRAM) Crossbar Array system, the researchers use a $64\times 64$ crossbar array, and each component is trainable, resulting in $4096$ parameters \cite{jung2022crossbar}.
The chemical autoencoder has $25$ independent BZ reaction chemical cells, which can be considered as $25$ parameters \cite{parrilla2020programmable}.
Finally, the phase change memory DNN has $156\times 156$ array \cite{yao2020fully}, and it holds $24336$ parameters. 

\begin{table}[h]
\caption{\textbf{| The number of trainable parameters shown in Fig.~\ref{fig:FIGINTRO}.} The numbers of parameters shown here are based on what the original papers claim or our reasonable estimate of the system from information they supply.}\label{tab_previous_research}
\begin{tabularx}{\linewidth}{@{}lllXr r@{}}
\toprule
\multicolumn{3}{c}{Architecture}   & Model & Year & Parameters \\
\midrule
\multicolumn{3}{c}{\multirow{10}{*}{Electronic Computers}} & AlexNet \cite{krizhevsky2012imagenet} & 2012 & 60,000,000 \\ 
\multicolumn{3}{l}{} & VGG-19 \cite{simonyan2014very} & 2014 & 144,000,000\\
\multicolumn{3}{l}{} & GPT-1 \cite{radford2018improving} & 2018 & 117,000,000\\
\multicolumn{3}{l}{} & BERT-Large \cite{devlin2019bert} & 2019 & 340,000,000\\
\multicolumn{3}{l}{} & Megatron-LM \cite{shoeybi2019megatron} & 2019 & 8,300,000,000\\
\multicolumn{3}{l}{} & GPT-2 \cite{radford2019language} & 2020 & 1,500,000,000\\
\multicolumn{3}{l}{} & T5-11B \cite{raffel2020exploring}  & 2020 & 11,000,000,000\\
\multicolumn{3}{l}{} & GPT-3 \cite{brown2020language} & 2020 & 175,000,000,000\\
\multicolumn{3}{l}{} & Switch Transformer \cite{fedus2022switch} & 2022 & 1,600,000,000,000\\
\multicolumn{3}{l}{} & PaLM \cite{chowdhery2023palm} & 2023 & 540,000,000,000\\
\midrule
\multirow{16}{*}{PNN} & \multirow{9}{*}{Optics}  & \multirow{3}{*}{Free Space Optics}  & Photonic RC \cite{bueno2018reinforcement} & 2018 & 2,025 \\
& & & D$^2$NN \cite{lin2018all} & 2018 & 450,000\\
& & & PAT \cite{wright2022deep} & 2022 & 50,568\\ 
& & & Reconf. ONN \cite{xia2023hardware} & 2023 & 200,000\\ 
& & & Analog Optical Computer \cite{kalinin2025analog} & 2025 & 5,000 \\ 
\cmidrule(lr){3-6}
& & Fiber Optics  & Optoelectric RC \cite{appeltant2011information} & 2011 & 400 \\ 
\cmidrule(lr){3-6}
& & \multirow{4}{*}{Photonics on Chip} & Integrated Photonic RC \cite{vandoorne2014experimental} & 2014 & 16\\ 
& & & Nanophotonic Neural Network \cite{shen2017deep} & 2017 & 112 \\
& & & Micro-Ring Resonator (MRR) RNN \cite{tait2017neuromorphic} & 2017 & 576 \\
& & & Mars \cite{ramey2020silicon} & 2020 & 4,096 \\
& & & 2D Programmable Waveguide \cite{onodera2025arbitrary} & 2025 & 10,000 \\
\cmidrule(lr){2-6}
& \multicolumn{2}{c}{\multirow{3}{*}{Hybrid}}  & DOPO Network \cite{inagaki2021collective} & 2021 & 57,600 \\
& \multicolumn{2}{l}{}  & Grad-free Deep RC \cite{nakajima2022physical} & 2022 & 810,424 \\
\cmidrule(lr){2-6}
& \multicolumn{2}{c}{\multirow{3}{*}{Spintronics}}  & Spin-Torque Oscillator RC \cite{torrejon2017neuromorphic} & 2017 & 400 \\
& \multicolumn{2}{l}{}  & Coupled Spin Oscillator Network \cite{romera2018vowel} & 2018 & 30 \\
& \multicolumn{2}{l}{}  & MRAM Crossbar Array \cite{jung2022crossbar} & 2022 & 4,096 \\
\cmidrule(lr){2-6}
& \multicolumn{2}{c}{MEMS}  & MEMS Nonlinear Resonator RC \cite{sun2021novel}  & 2021 & 440 \\
\cmidrule(lr){2-6}
& \multicolumn{2}{c}{\multirow{2}{*}{Chemitronics}}  & Chemical Autoencoder \cite{parrilla2020programmable} & 2020 & 25 \\
& \multicolumn{2}{l}{}  & Electrochemical RC \cite{kan2022physical} & 2022 & 112 \\
\cmidrule(lr){2-6}
& \multicolumn{2}{c}{Memristor} & Phase-Change Memory DNN \cite{yao2020fully} & 2020 & 24,336 \\
\botrule
\end{tabularx}
\end{table}

\section{Parameter-scaling comparison with representative models}
In the main text, we showed that TIDAL-Net improves performance on the SVHN image-classification and Shakespeare next-token prediction tasks as the number of distinct weight banks $L_\mathrm{W}$ increases, while the number of temporal steps $L_\mathrm{T}$ is kept fixed.
However, the raw changes in error rate alone do not immediately indicate how meaningful these gains are in terms of model size.
To provide this context, we compare TIDAL-Net with conventional architectures on the same task from the viewpoint of parameter scaling.

\subsection{Simple CNN}
To contextualize the image-classification performance of TIDAL-Net, we trained additional reference models based on conventional digital convolutional neural networks. 
These models were used to compare TIDAL-Net against standard digital convolutional architectures with a similar order of trainable parameters. 
Unlike TIDAL-Net, these CNNs do not use temporal weight reuse. Instead, they exploit the standard spatial weight sharing of convolutional kernels.

As reference models, we considered a family of plain convolutional neural networks, which we refer to as Simple CNNs.
This model family was designed to be scalable through two hyperparameters: the list of channel widths in the convolutional stages and the width of the intermediate fully connected layer.
This allowed us to construct a family of models with different trainable parameter counts while keeping the overall architectural principle fixed.

The Simple CNN baseline consists of a sequence of convolutional stages followed by a fully connected classifier. 
Each convolution stage is composed of a two-dimensional convolution, batch normalization, ReLU nonlinearity, and $2\times2$ max pooling. 
For an input feature map $u_l$, the convolutional part of one stage can be written as
\begin{align}
u_{l+1}&=\mathrm{MaxPool}_{2\times2}[\phi (\mathrm{BN} (W_l*u_l) )]
\end{align}
where $W_l$ is a $3\times3$ convolutional kernel, $*$ denotes convolution, $\mathrm{BN}$ denotes batch normalization, and $\phi(\cdot)$ is the ReLU activation function.
The convolutional layers do not use bias terms because they are followed by batch normalization.

After the final convolutional stage, the feature tensor is flattened and passed through a two-layer fully connected classifier,
\begin{align}
z&=W_2\phi(W_1\mathrm{vec}(u_L)+b_1)+b_2
\end{align}
where $z\in\mathbb{R}^{10}$ is the output logit vector for the ten-class classification task. 
Dropout with probability $0.5$ is applied between the two fully connected layers.

For a $32\times32$ input image, if the Simple CNN model has $K$ convolutional stages, the spatial resolution after the convolutional feature extractor is $32/2^K$.

For each CNN baseline, the model was trained as a standard supervised image classifier. Given an input image $x_i$, the network outputs logits 
\begin{align}
z_i&=F_{\theta}(x_i),
\end{align}
where $F_{\theta}$ denotes the Simple CNN model. 
The predicted label is
\begin{align}
\hat{y}_i&=\arg \max_c z_{i,c}.
\end{align}
The loss function was the cross-entropy loss,
\begin{align}
L&=-\frac{1}{B}\sum_{i=1}^{B}\log{\frac{\exp{(z_{i,y_i})}}{\sum_{c=1}^{10}\exp{(z_{i,c})}}}
\end{align}
where $B$ is the mini-batch size and $y_i$ is the ground-truth class label.
The reported classification error was
\begin{align}
\mathrm{Error\ Rate (\%)}&=100\left[1-\frac{1}{N}\sum_{i=1}^{N}\mathbf{1}(\hat{y}_i=y_i)\right]
\end{align}

The CNN baselines were trained using the same training script as the other reference CNN experiments. 
The optimizer was Adam, and the training script recorded the training, validation, and test losses and error rates at each epoch. 
The final result files also stored the model name, channel configuration, fully connected dimension, number of trainable parameters, dataset name, learning rate, weight decay, random seed, augmentation setting, and noise level.

\subsection{Comparison}
\begin{figure}[h]
\centering
\includegraphics[width=0.7\textwidth]{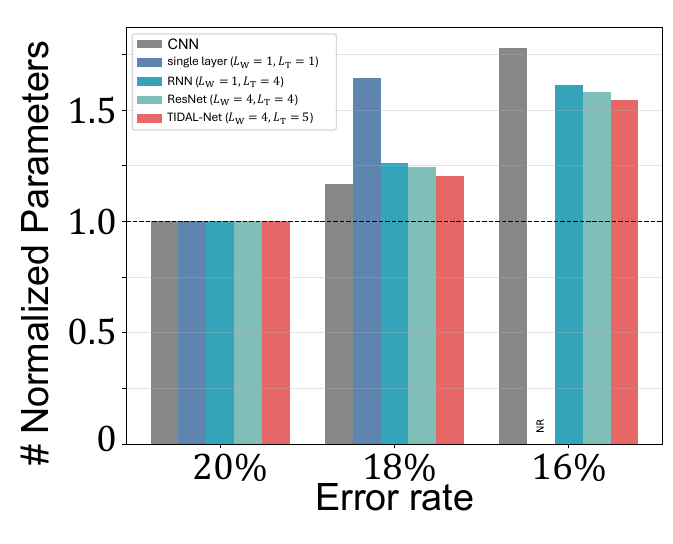}
\caption{\textbf{|Parameter scaling comparison between CNN and TIDAL-Net.} Normalized number of trainable parameters required to reach target error rates of $20\%$, $18\%$, and $16\%$ on image classification.
For each model family, the required parameter count is normalized by the parameter count required to reach 20\% error.
This comparison indicates that TIDAL-Net achieves a competitive accuracy--parameter trade-off while using a temporal weight-reuse mechanism.
``NR'' denotes ``not reached'', indicating that the model did not reach the corresponding target error rate within the tested parameter range.
}\label{fig:param_scale}
\end{figure}

Figure~\ref{fig:param_scale} compares the number of parameters required to reach several target error rates on SVHN classification performance of CNN, RNN, and TIDAL-Net. 
Specifically, the parameter counts required to achieve $16\%$,$18\%$, and $20\%$ error are normalized by the parameter count required at $20\%$ error.
The required parameter counts were estimated by linear interpolation between neighboring points in the error--parameter curves.

This comparison shows that the parameter scaling of CNNs and TIDAL-Net is broadly similar on this task. 
In other words, the gains obtained by TIDAL-Net are not merely visible in raw error values, but correspond to a competitive accuracy–parameter trade-off. 
In general, we see that to reduce the testing error rate on SVHN from $20\%$ to $18\%$ requires an increase of trainable parameters of around $1.2$ or $1.3$-times compared to the baseline. 
Meanwhile, in the main manuscript, it is shown that the TIDAL-Net improves error rates on SVHN from around $19\%$ to around $17\%$ without increasing the number of trainable parameters. 
Thus, the reuse of layers leads to a significant increase in terms of parameter efficiency (although not by orders-of-magnitude - more improvements of PNNs will be needed besides clever time-multiplexing).

At the same time, Figure~\ref{fig:param_scale} should not be interpreted as indicating that TIDAL-Net universally performs as good or better than CNNs. 
Rather, it shows that TIDAL-Net can achieve comparable scaling behavior while relying on a weight-reuse mechanism that is motivated by hardware constraints in PNNs.
Due the significantly more efficient parameter encoding of CNNs, by using Kernel functions, the overall number of parameters needed for a particular performance is lower in CNNs, when compared to DNNs, TIDAL-Net or RNNs.

Importantly, TIDAL-Net is compatible with convolutional layers (or other types of layers) in principle. 
TIDAL-Net relies on quickly switching between existing layers, but does in principle not prescribe the nature of these layers.
A TIDAL-CNN could time-multiplex convolutional layers (or a mix of convolutional and fully connected layers), according to the needs of the task and the available hardware on a specific PNN platform. 
Such mixed-layer time-multiplexed PNNs remain promising research objects for the future.


\section{Timescales of parameter training in PNNs}\label{section_timescale}
As discussed in the main manuscript, the advantage of TIDAL-Net arises from fast switching among preconfigured weight banks, rather than from fast reconfiguration of the trainable physical parameters themselves.
We also discussed the requirements for implementing TIDAL-Net and some architecture examples on MZI networks.
In this section, we provide additional information on the reconfiguration time and clarify the advantage of TIDAL-Net for existing PNN architectures.

TIDAL-Net is a framework that exploits the time-scale separation between fast inference dynamics and slow full-network reconfiguration via switching.
This timescale separation can arise either because full-network reconfiguration becomes slower as the number of tunable parameters increases, or because switching and weight tuning rely on different physical mechanisms.
Thus, the relevance of TIDAL-Net should be assessed for each target platform by comparing full-network reconfiguration with switching and other inference latencies.

From an architectural point of view, the inference time should be estimated by decomposing the computation into the individual operations performed in the system.
As shown in Fig.~\ref{fig:relax_net_PNN}, a key advantage of TIDAL-Net is that it enables computation with larger effective networks with the same hardware.
It is therefore natural to analyze systems such as time-variable RNNs and TIDAL-Net under the assumption that as many operations as possible are executed in parallel.
For example, in MZI-based systems, network reconfiguration and nonlinear activation may be performed simultaneously if they are implemented using different physical components~\cite{shen2017deep}.

To compare full reconfiguration and switching, we model one recurrent step as a coarse-grained scheduling process. First, the selected MZI weight bank performs the linear matrix multiplication. After the output of this MZI block becomes available to the subsequent state-update block, two operations can proceed in parallel: the remaining state update and the preparation of the next weight bank. The next recurrent step can start only after both operations have finished. Therefore, the latency after the MZI matrix multiplication is determined by the slower of these two operations.

Let $T_{\mathrm{MM}}$ denote the latency of the MZI-based matrix-multiplication block, including the time required for its output to become available to the subsequent state-update block. Let $T_{\mathrm{state}}$ denote the latency of the state-update block, including nonlinear activation, residual combination, and regeneration of the next hidden state. With this notation, the per-step latency of a time-variable RNN is approximated as
\begin{align}
T_{\mathrm{step}}^{\mathrm{time-variable\ RNN}}
&=
T_{\mathrm{MM}}
+
\max\left(T_{\mathrm{upd}},T_{\mathrm{state}}\right),
\label{eq:time:step_time_variable_RNN}
\end{align}
where $T_{\mathrm{upd}}$ is the time required for full-network reconfiguration. 
By contrast, TIDAL-Net only requires switching among preconfigured weight banks, giving
\begin{align}
T_{\mathrm{step}}^{\mathrm{TIDAL-Net}}
&=
T_{\mathrm{MM}}
+
\max\left(T_{\mathrm{swt}},T_{\mathrm{state}}\right),
\label{eq:time:step_TIDAL_Net}
\end{align}
where $T_{\mathrm{swt}}$ is the time required for switching among the preconfigured weight banks.

The total inference time of a time-variable RNN is then approximated as
\begin{align}
T_{\mathrm{infer}}^{\mathrm{time-variable\ RNN}}
&= T_{\mathrm{init}} + N_{\mathrm{step}}T_{\mathrm{step}}^{\mathrm{time-variable\ RNN}} + T_{\mathrm{out}},
\label{eq:time:time-variable_RNN}
\end{align}
  whereas the total inference time of TIDAL-Net is
\begin{align}
T_{\mathrm{infer}}^{\mathrm{TIDAL-Net}}&= T_{\mathrm{init}}
+ N_{\mathrm{step}}T_{\mathrm{step}}^{\mathrm{TIDAL-Net}}
+ T_{\mathrm{out}}.
  \label{eq:time:TIDAL-Net}
\end{align}
Equivalently, the difference between the two models is that the full reconfiguration time $T_{\mathrm{upd}}$ in the time-variable RNN is replaced by the switching time $T_{\mathrm{swt}}$ in TIDAL-Net. 
These expressions are intended as a coarse-grained latency model for physical recurrent systems in which state update and weight-bank preparation can be at least partially overlapped. 
Thus, TIDAL-Net is advantageous when $T_{\mathrm{swt}}\ll T_{\mathrm{upd}}$ and the other latencies do not dominate the total inference time. The actual values of $T_{\mathrm{infer}}^{\mathrm{TIDAL-Net}}$ and $T_{\mathrm{infer}}^{\mathrm{time-variable\ RNN}}$ depend on the device characteristics of the components used in the target system.



One promising platform for implementing TIDAL-Net is a recurrent physical system built around a coherent linear processor based on MZIs.
In this sense, TIDAL-Net may be naturally realized by extending previous coherent photonic neural-network and recurrent/reservoir-computing architectures.
Representative examples include the optical neural network architecture~\cite{shen2017deep} and the photonic reservoir-computing architecture~\cite{nakajima2021scalable}.
In both cases, the central linear operation is implemented using an MZI-based coherent processor.
In the optical neural network architecture ~\cite{shen2017deep}, the linear layer is formulated as an Optical Interference Unit (OIU), whereas the photonic reservoir computing employs an integrated coherent linear photonic processor for both the input and recurrent linear transformations~\cite{nakajima2021scalable}.
These MZI-based processors are naturally combined with photodetection and feedback paths to realize nonlinear and recurrent dynamics, which is an opto-electrical-optical conversion.
In the proof-of-concept experiment of the optical neural network, the nonlinear transformation was implemented in the electronic domain after photodetection, and the resulting signal was then injected into the next optical stage~\cite{shen2017deep}.
By contrast, the photonic reservoir computer presents a coherent photonic reservoir computing system in which both the input and recurrent weights are encoded in the spatiotemporal domain through photonic linear processing~\cite{nakajima2021scalable}.

In addition to MZI circuits, another key component of photonic computing systems is the Optical–Electrical–Optical (OEO) converter, which provides reconfigurable nonlinear transfer, signal regeneration, and gain.
It can therefore serve as a practical nonlinear element or feedback interface in recurrent photonic neural networks.
Such an optoelectronic building block can be realized, for example, using monolithically integrated silicon-photonic OEO converters in either load-resistor or current-injection configurations, each combining a germanium photodetector with a microring modulator~\cite{arahata2026silicon}.

In such MZI-based photonic recurrent neural-network systems, the evolution equation can be written, at a coarse-grained level, as
\begin{align}
\tau\frac{d\mathbf{u}}{dt}&=-\mathbf{u}+\Phi(W_{\sigma(t)}\mathbf{u}+W_{\mathrm{in}}\mathbf{x}(t)+b_{\sigma(t)})\label{eq:evolve}
\end{align}
where $\tau$ is the effective response time of the hidden state dynamics.
Here, $\mathbf{u}$ denotes the internal state of the system, such as an electrical signal obtained after photodetection.
$W_{\sigma(t)}$ is the selected linear weight bank at time $t$, and $b_{\sigma(t)}$ is the corresponding bias term.
The switching index $\sigma(t)$ selects one element from a finite set of preconfigured matrices, $\{W_1,\ldots,W_{L_{\mathrm{W}}}\}$.
The function $\Phi(\cdot)$ represents the nonlinear activation transformation implemented by an optoelectronic nonlinear element, such as an OEO converter and photo-detector.

When we discretize the system by defining $h[n]:=\mathbf{u}(n\Delta T)$, $x[n]:=x(n\Delta T)$, $\sigma_n:=\sigma(n\Delta T)$ and coarse-grain the dynamics over each interval $[n\Delta T,(n+1)\Delta T)$ under the approximation that the argument of the nonlinear function remains constant within the interval, we obtain
\begin{align}
h[n+1]&=\rho h[n]+(1-\rho)\Phi(W_{\sigma(t)}h[n]+W_{\mathrm{in}}x[n]+b_{\sigma(t)})\label{eq:rho}
\end{align}
with
\begin{align}
\rho&:=e^{-\Delta T/\tau}
\end{align}
By identifying $\alpha=\rho$ and
$f(\cdot)=(1-\rho)\Phi(\cdot)$, Eq.~\ref{eq:rho} can be written in the same form as Eq.~\ref{eq:relax_net}.
In other words, if the physical system can be described by Eq.~\ref{eq:evolve}, which may be viewed as a neural-ODE-like continuous-time model, then Eq.~\ref{eq:relax_net} can be interpreted as its time-discretized implementation.
This provides a useful guideline for designing TIDAL-Net using physical devices.

Based on these considerations, we now compare
$T_{\mathrm{infer}}^{\mathrm{time-variable\ RNN}}$ and $T_{\mathrm{infer}}^{\mathrm{TIDAL-Net}}$
to assess whether TIDAL-Net can provide a computing time advantage in MZI-based networks.
Table~\ref{tab_component} summarizes representative timescales for the individual components involved in information processing in MZI-based photonic computing systems.
The representative timescales in Table~\ref{tab_component} indicate that thermo-optic full-network reconfiguration is typically much slower than optical matrix multiplication, nonlinear activation, feedback, and fast switching. 
Thus, when $T_{\mathrm{swt}} \ll T_{\mathrm{upd}}$ and the other latencies are not dominant, Eq.~\eqref{eq:time:TIDAL-Net} and Eq.~\eqref{eq:time:time-variable_RNN} predict a computing time advantage for TIDAL-Net over a time-variable RNN, $T_{\mathrm{infer}}^{\mathrm{time-variable\ RNN}}\gg T_{\mathrm{infer}}^{\mathrm{TIDAL-Net}}$.

\begin{table}[h]
\caption{\textbf{Computing-time components for an MZI-based photonic recurrent system.}\\
Representative latency components used in the coarse-grained timing model.\\
Here, $T_{\mathrm{upd}}$ is the latency of full MZI-weight reconfiguration, $T_{\mathrm{swt}}$ is the latency of switching among preconfigured weight banks, $T_{\mathrm{MM}}$ is the latency of the MZI-based matrix-multiplication block, and $T_{\mathrm{state}}$ is the latency of the subsequent state-update block.\\
The listed timescales are representative values and depend on the device implementation~\cite{duan2021low,hornbeck1997digital,friberg1988femotosecond,li2018ultra,shen2017deep,arahata2026silicon}.}
\label{tab_component}%
\small
\setlength{\tabcolsep}{4pt}
\begin{tabularx}{\textwidth}{@{}l l X l@{}}
\toprule
Time & Component & Main limiting process & Timescale \\
\midrule
$T_{\mathrm{upd}}$
& MZI phase shifters
& Full weight reconfiguration, e.g., thermo-optic tuning
& $\mathrm{\mu s}$--$10\,\mathrm{\mu s}$ or longer \\
$T_{\mathrm{swt}}$
& Optical/electrical switch
& Weight-bank selection
& sub-$\mathrm{ns}$--$\mathrm{\mu s}$ \\
$T_{\mathrm{MM}}$
& MZI mesh
& Optical matrix multiplication through the interferometer mesh
& $\mathrm{ps}$-scale \\
$T_{\mathrm{state}}$
& State-update block
& Photodetection, amplification, modulation, nonlinear activation, and optical combining
& $\mathrm{ps}$--$10\,\mathrm{ns}$ \\
\botrule
\end{tabularx}
\end{table}

Another promising TIDAL-Net application target is a Micro-Ring Resonator (MRR) weight bank system \cite{tait2017neuromorphic}.
In that system, the parameters of the MRR are changed by adjusting the temperature of its components \cite{tait2018feedback}.
Depending on conditions (e.g., required accuracy), weight reconfiguration using thermo-optic effects typically requires sub-microseconds \cite{bogaerts2012silicon}, which is slow compared to the fast dynamic timescales of optical setups \cite{hornbeck1997digital,friberg1988femotosecond}.
Because such systems can be combined with optoelectronic conversion and fast switching components, the timescale separation between slow weight reconfiguration and faster inference/switching operations suggests that TIDAL-Net could be beneficial for this platform.
In other words, when applying this system to a time-variable neural network, the computation speed is dominated by the thermo-optic effect, which TIDAL-Net can avoid by switching.

PNNs using Magnetic Tunnel Junction (MTJ) crossbar arrays \cite{jung2022crossbar} also represent a potential platform for TIDAL-Net.
In their $64\times64$ MTJ system, the total reconfiguration time for the entire matrix is on the order of tens of microseconds.
Considering other electrical system components can work faster than magnetic reconfiguration, by incorporating switching in its electronic components, TIDAL-Net is also applicable to non-photonic PNNs and helps it achieve a time-variable Recurrent Neural Network (RNN).

More generally, TIDAL-Net is most relevant for PNN platforms in which the trainable parameter-setting mechanism is slower than the forward inference or switching dynamics.
In such systems, adding fast switching or routing components can allow the network to cycle through a small set of preconfigured weight banks without reprogramming the physical parameters at each time step.
This modest hardware requirement constitutes one of the key advantages and defining features of the TIDAL-Net framework.


\end{appendices}


\end{document}